\documentclass[letterpaper]{article} 
\usepackage{aaai25}  
\usepackage{times}  
\usepackage{helvet}  
\usepackage{courier}  
\usepackage[hyphens]{url}  
\usepackage{graphicx} 
\usepackage{natbib}  
\usepackage{caption} 
\usepackage{algorithm}
\usepackage{algorithmic}
\usepackage{subfigure}
\usepackage{multirow}
\usepackage{array}
\usepackage{booktabs}
\usepackage{amsmath,amsfonts}
\usepackage{xcolor}
\usepackage{hyperref}
\usepackage{newfloat}
\usepackage{listings}
\DeclareCaptionStyle{ruled}{labelfont=normalfont,labelsep=colon,strut=off} 
\floatstyle{ruled}
\newfloat{listing}{tb}{lst}{}
\floatname{listing}{Listing}
\title{Distribution-Consistency-Guided Multi-modal Hashing}

\author {
    Jin-Yu Liu,
    Xian-Ling Mao\thanks{Corresponding author.},
    Tian-Yi Che,
    Rong-Cheng Tu
}
\affiliations {
    School of Computer Science and Technology, Beijing Institute of Technology, Beijing, China\\
    \{liujinyu1229, turongcheng\}@gmail.com,
    \{maoxl, ccty\}@bit.edu.cn
}

\usepackage{bibentry}

\begin{document}

\maketitle

\begin{abstract}
Multi-modal hashing methods have gained popularity due to their fast speed and low storage requirements. Among them, the supervised methods demonstrate better performance by utilizing labels as supervisory signals compared with unsupervised methods. Currently, for almost all supervised multi-modal hashing methods, there is a hidden assumption that training sets have no noisy labels. However, labels are often annotated incorrectly due to manual labeling in real-world scenarios, which will greatly harm the retrieval performance. To address this issue, we first discover a significant distribution consistency pattern through experiments, i.e., the 1-0 distribution of the presence or absence of each category in the label is consistent with the high-low distribution of similarity scores of the hash codes relative to category centers. Then, inspired by this pattern, we propose a novel \textbf{D}istribution-\textbf{C}onsistency-\textbf{G}uided \textbf{M}ulti-modal \textbf{H}ashing (\textbf{DCGMH}), which aims to filter and reconstruct noisy labels to enhance retrieval performance. Specifically, the proposed method first randomly initializes several category centers, each representing the region's centroid of its respective category, which are used to compute the high-low distribution of similarity scores; Noisy and clean labels are then separately filtered out via the discovered distribution consistency pattern to mitigate the impact of noisy labels; Subsequently, a correction strategy, which is indirectly designed via the distribution consistency pattern, is applied to the filtered noisy labels, correcting high-confidence ones while treating low-confidence ones as unlabeled for unsupervised learning, thereby further enhancing the model’s performance. Extensive experiments on three widely used datasets demonstrate the superiority of the proposed method compared to state-of-the-art baselines in multi-modal retrieval tasks. The code is available at \href{https://github.com/LiuJinyu1229/DCGMH}{https://github.com/LiuJinyu1229/DCGMH}.
\end{abstract}

%

\section{Introduction}

With the rapid growth of multimedia data such as images, text, and videos, achieving effective retrieval from massive multi-modal data has become a significant challenge. To address this challenge, numerous information retrieval technologies have emerged \cite{jiang2017deep, sung2018learning, zhang2018attention, chen2020uniter}, with multi-modal hashing methods \cite{liu2014multiple, chen2020enhanced, zhu2021efficient, lu2021graph, wu2022online} gaining widespread attention for their fast retrieval speed and low storage requirements. Unlike uni-modal hashing \cite{tu2018object, tu2021weighted, guo2022intra, tu2020mls3rduh, tu2021partial} and cross-modal hashing \cite{zhang2018unsupervised, tu2022deep, tu2023unsupervised, tu2021hashing}, multi-modal hashing maps data points from different modalities into a unified Hamming space for fusion, resulting in binary hash codes that facilitate efficient multi-multi retrieval. Compared to unsupervised methods \cite{song2013effective, shen2015multi, shen2018multiview, zheng2020efficient, wu2021multi}, supervised multi-modal hashing methods \cite{yang2017discrete, xie2017dynamic, yu2022hadamard, zheng2024lcemh} generate more discriminative hash codes and achieve more accurate retrieval by utilizing labels as supervisory signals and there is a hidden assumption that training sets have no noisy labels for these supervised methods.

However, in real-world scenarios, labels may be incorrectly annotated due to manual labeling, such as an image that should be labeled as "tiger" being mistakenly labeled as "cat", which limits the applicability of existing supervised hashing methods in noisy label scenarios. While some works \cite{sun2022heart, yang2022mutual} in image hashing and cross-modal hashing have demonstrated that the presence of noisy labels in the training set can lead to model overfitting, resulting in indistinguishable hash codes and inaccurate retrieval, no work has yet focused on and resolved this issue in the field of multi-modal hashing.

To effectively tackle this issue, it is crucial to filter out noisy labels from the dataset. Existing single or cross-modal hashing methods have demonstrated that models initially learn effective hash mappings from clean labels but eventually overfit to noisy labels, resulting in degraded performance. This phenomenon suggests that even after a short training period, the generated hash codes are somewhat discriminative and tend to align with their corresponding category centers \cite{yuan2020central, tu2023data}, as indicated by higher similarity scores. Thus, the 1-0 distribution of the category's presence or absence, i.e., the label vector, should be consistent with the high-low distribution of similarity scores between the hash codes and category centers. Specifically, if an instance belongs to a category, the corresponding label bit should be '1', and the hash code should show a higher similarity score with the category center; otherwise, the bit should be '0', and the similarity score should be lower. Therefore, we hypothesize that this pattern of distribution consistency can effectively filter out noisy labels.

\begin{figure}[]
	\centering
	\includegraphics[width=1.0\linewidth]{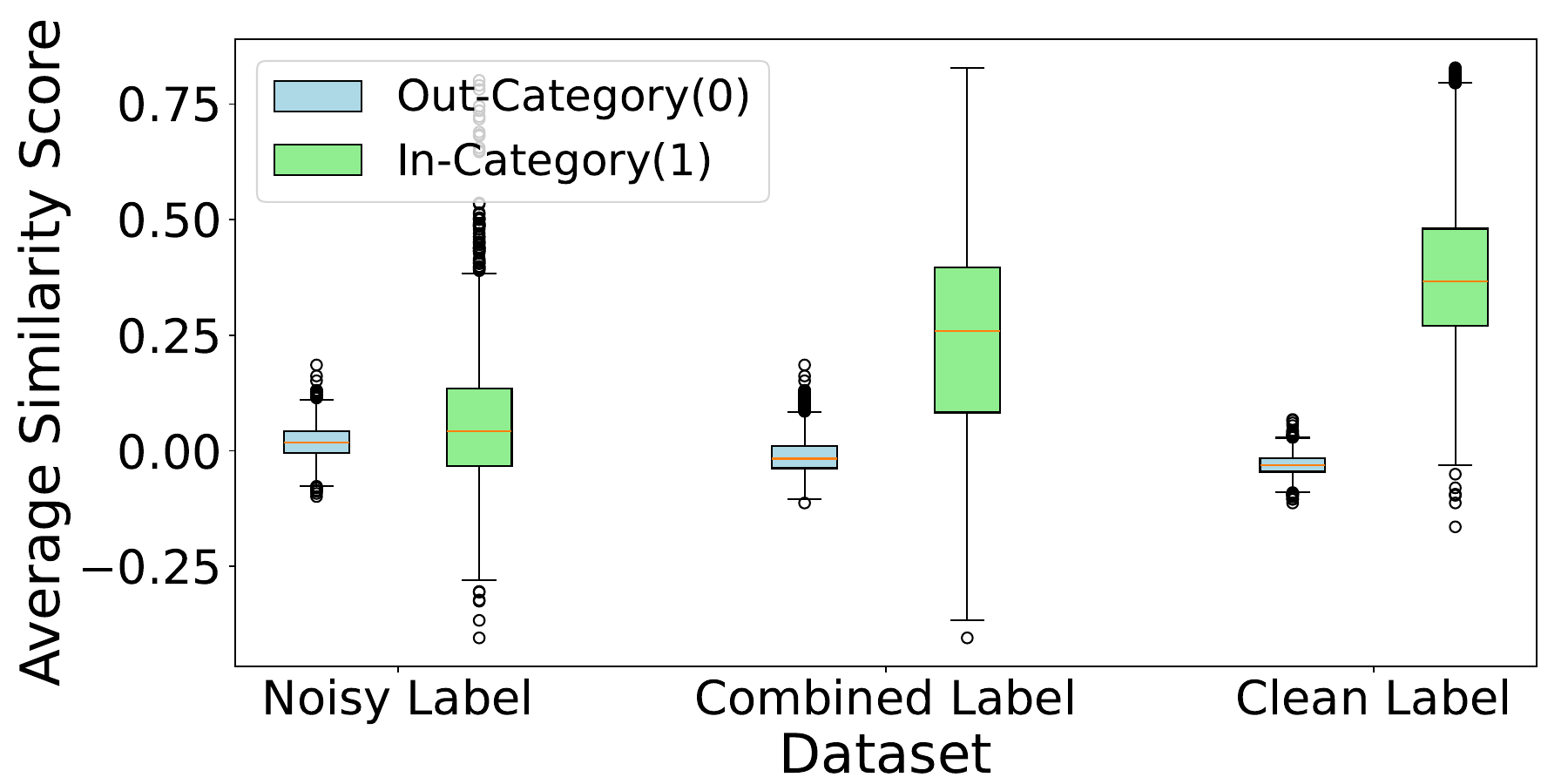}
	\caption{BoxPlot comparison of average similarity scores for in-category and out-category across clean and noisy label datasets, where "Out-Category(0)" represents the box plot distribution of the average similarity scores of hash codes to all categories it does not belong, while "In-Category(1)" represents the box plot distribution for the belonging categories, and the horizontal line within each box indicates the median of all average similarity scores.}
	\label{comparision}
\end{figure}

To validate this hypothesis, we conduct a BoxPlot statistical analysis comparing the average similarity scores between hash codes and their respective in-category and out-category centers. This analysis is performed on the MIR Flickr dataset which contains a training set of 5,000 instances with a noisy label ratio of 40\%. After training the model for 10 epochs, we separate the dataset into noisy and clean subsets and analyze the distribution of similarity scores. The results, shown in Figure \ref{comparision}, reveal that in the clean label dataset, the similarity scores of hash codes to their respective category centers are significantly higher compared to non-belonging centers. In contrast, in the noisy label dataset, this difference is less pronounced due to the misalignment between assigned categories and actual hash code semantics. This observation confirms that there is a notable consistency between the 1-0 label distribution and the high-low similarity score distribution in clean labels, which is disrupted in noisy labels. Therefore, by exploiting these consistency differences, we can effectively filter out noisy labels, supporting our hypothesis.

Consequently, inspired by the distribution consistency pattern and the hypothesis, we propose a novel \textbf{D}istribution-\textbf{C}onsistency-\textbf{G}uided \textbf{M}ulti-modal \textbf{H}ashing (\textbf{DCGMH}) via the consistency between the 1-0 distribution of labels and the high-low distribution of similarity scores to filter and reutilize noisy labels, thereby preventing the model from overfitting to noisy labels and improving its retrieval performance in real-world scenarios. Specifically, the proposed method first randomly initializes several category centers to compute the similarity scores of the hash codes relative to each category center, with each center representing the central region of its corresponding category; Then, based on the distribution consistency pattern, the noisy and clean labels are separately filtered out to mitigate the impact of noisy labels; Subsequently, for the noisy label set, we design a reconstruction strategy via the distribution consistency pattern to correct high-confidence noisy labels, while low-confidence noisy label instances are treated as unlabeled instances to facilitate unsupervised learning for extracting semantic information and further enhancing the performance of the multi-modal hashing model. In conclusion, the main contributions of the proposed method are as follows:
\begin{itemize}
    \item We discover the consistency between the 1-0 distribution of the presence or absence of each category in the label and the high-low distribution of similarity scores of the hash codes relative to each category center and validate it through BoxPlot statistical analysis.
    \item We design a filter via the distribution consistency pattern to filter out noisy labels and improve the applicability of supervised multi-modal hashing methods in real-world scenarios. To the best of our knowledge, no similar work has been done.
    \item We design a corrector via the distribution consistency pattern to correct high-confidence noisy labels and utilize low-confidence ones for unsupervised learning.
    \item Extensive experiments on three benchmark datasets demonstrate the proposed method outperforms the state-of-the-art baselines in multi-modal retrieval tasks.
\end{itemize}

\section{Related Work}

\subsection{Multi-modal Hashing}

In multi-modal hashing, supervised methods utilize labels as supervisory information to help models learn richer semantic information, and can be categorized into shallow and deep approaches based on whether deep networks are utilized. Shallow supervised multi-modal hashing \cite{lu2019flexible, lu2019efficient, zheng2019fast, zheng2022efficient, an2022cognitive} often relies on linear mapping or matrix factorization to model the latent semantic associations between modalities. For instance, OMHDQ \cite{lu2019online} links hash code learning with both low-level data distribution and high-level semantic distribution based on paired semantic labels. SAPMH \cite{zheng2020adaptive} employs paired semantic labels as parameter-free supervisory information to learn multidimensional latent representations. In contrast, deep supervised multi-modal hashing \cite{yan2020deep, zhu2020deep, shen2023graph, lu2020semantic, tan2023partial} leverages deep networks to integrate feature extraction and hash code learning into a unified deep framework. For example, BSTH \cite{tan2022bit} introduces a bit-aware semantic transformation module to achieve fine-grained, concept-level alignment and fusion of multi-modal data, thereby generating high-quality hash codes. STBMH \cite{tu2024similarity} addresses the issue of similarity transitivity broken in multi-label scenarios by designing an additional regularization term.

For almost all supervised multi-modal hashing methods, there is a hidden assumption that training sets have no noisy labels. However, in real-world scenarios, the labels are often annotated incorrectly due to manual labeling which greatly harms the model's performance. This issue necessitates effective strategies for noisy label learning to maintain model robustness and accuracy.

\subsection{Noisy Label Learning}
Noisy label learning has been extensively studied in tasks such as image classification. Existing approaches to handling noisy labels can be broadly categorized into two types: noise-robust modeling \cite{song2020convex, song2022learning, shu2019meta} and label-noise cleaning \cite{zheng2020error, wu2021learning, kim2021fine, wei2022self}. Noise-robust modeling involves directly training robust models on noisy labels using noise-specific loss functions or regularization terms; for instance, NCR \cite{iscen2022learning} designs a regularization loss term based on the consistency between an instance and its neighboring nodes in the feature space. In contrast, label-noise cleaning focuses on filtering or correcting noisy labels directly, exemplified by FCF \cite{jiang2024more}, which introduces a fusion cleaning framework that combines correction and filtering to address different types of noisy labels.

In the context of hashing, there have been works addressing noisy labels in image hashing \cite{sun2022heart, wang2023dior} and cross-modal hashing \cite{yang2022mutual, li2024robust} by using loss-based methods. For instance, DIOR \cite{wang2023dior} uses the concept of behavior similarity between original and augmented views to filter noisy labels, while CMMQ \cite{yang2022mutual} designs a proxy-based contrastive loss to mitigate the impact of noisy labels. However, no existing work addresses the issue of noisy labels in multi-modal hashing retrieval. Based on the discovered distribution consistency pattern, we propose a novel approach called DCGMH to filter and reutilize noisy labels, improving the robustness of the hashing model. The specific details of this approach will be discussed in the next section.

\section{The Proposed Method}

\begin{figure*}[]
	\centering
	\includegraphics[width=1\linewidth]{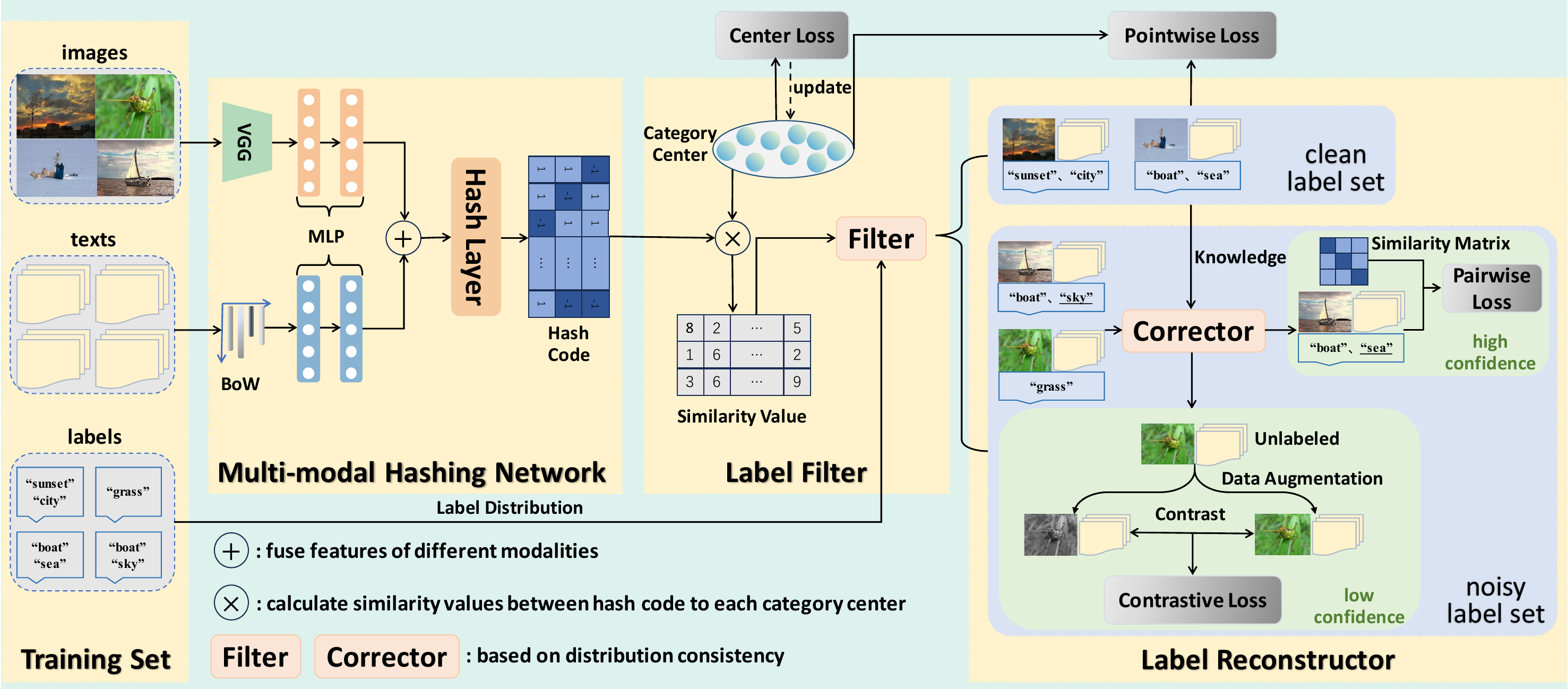}
	\caption{The architecture of our proposed DCGMH.}
	\label{architecture}
\end{figure*}

In the section, we first describe the problem definition and then provide detailed explanations of our proposed method's architecture, whose framework diagram is depicted in Figure \ref{architecture}. Subsequently, we summarize the objective function and optimize the model's training process. Finally, we demonstrate the out-of-sample extension.

\subsection{Problem Definition}
Similar to most existing multi-modal hashing methods, this work focuses on image-text datasets. Assuming that there is a dataset $\boldsymbol{O}$ containing $n$ instances, denoted as $\boldsymbol{O} = \{\boldsymbol{o}_i\}_{i=1}^n = \{\boldsymbol{x}_i, \boldsymbol{y}_i, \boldsymbol{l}_i\}_{i=1}^n$, where $\boldsymbol{x}_i$ and $\boldsymbol{y}_i$ represents the text and image modal data point, respectively. Moreover, $\boldsymbol{l}_i \in \{0, 1\}^{m}$ represents the label vector of the instance $\boldsymbol{o}_i$, where $m$ is the total number of categories, and when an instance $\boldsymbol{o}_i$ belongs to the category $j$, $\boldsymbol{l}_{ij} = 1$; otherwise, $\boldsymbol{l}_{ij} = 0$. Furthermore, the similarity matrix $\boldsymbol{S} \in \{-1, 1\}^{n \times n}$ is employed to represent the similarity between instances, such that when instances $\boldsymbol{o}_i$ and $\boldsymbol{o}_j$ share at least one category, $\boldsymbol{s}_{ij} = 1$, indicating they are similar; otherwise, $\boldsymbol{s}_{ij} = -1$, indicating they are dissimilar. Additionally, the instances in dataset $\boldsymbol{O}$ are ultimately mapped to hash codes $\boldsymbol{B} \in \{-1, 1\}^{n \times k}$ in the Hamming space while preserving the original semantics, where $k$ represents the length of the hash codes.

\subsection{Architecture}

\subsubsection{Multi-modal Hashing Network}
To generate high-quality hash codes, we use instance $\boldsymbol{o}_i = \{\boldsymbol{x}_i, \boldsymbol{y}_i, \boldsymbol{l}_i\}$ as the training data for the hashing network. Then, we employ a bag-of-words (BoW) \cite{simonyan2014very} model to extract the feature representation $\boldsymbol{f}_i^x$ and a VGG \cite{simonyan2014very} model without the final classification layer to extract the feature representation $\boldsymbol{f}_i^y$ for the text and image modal data, respectively. Subsequently, modality-specific multi-layer perceptions (MLP) are utilized to map the feature representation $\boldsymbol{f}_i^{*}, * \in \{x, y\}$ of each modality to a unified space, denoted as:
\begin{equation}
    \label{mlp}
    \begin{aligned}
        \boldsymbol{u}_i^{*} = \boldsymbol{MLP}^{*}(\boldsymbol{f}_i^{*};\theta^{*})
    \end{aligned}
\end{equation}
where $\boldsymbol{u}_i^{*}$ is the resulting projected feature representation and $\theta^{*}$ denotes the learnable parameters of the MLP for the respective modality. Finally, the projected features from different modalities are fused by directly summing them, and this fused representation is then passed through a hashing function with a non-linear activation function ($tanh(\cdot)$) to generate the fused hash code $\boldsymbol{\hat{b}}_i$, denoted as:
\begin{equation}
    \label{hash}
    \begin{aligned}
        \boldsymbol{\hat{b}}_i = \mathcal{H}(\boldsymbol{u}_i^x + \boldsymbol{u}_i^y;\theta)
    \end{aligned}
\end{equation}
where $\mathcal{H}$ refers to the hash function and $\theta$ is a set of learnable parameters. 
Finally, the final binary hash code $\boldsymbol{b}_i$ for instance $\boldsymbol{o}_i$ can be obtained as $\boldsymbol{b}_i = sgn(\boldsymbol{\hat{b}}_i)$, where $sgn(\cdot)$ is a function that maps positive values to 1 and negative values to -1. 
In summary, for instance $\boldsymbol{o}_i$, its binary hash code can be formulated as $\boldsymbol{b}_i = sgn(\mathcal{F}(\boldsymbol{x}_i, \boldsymbol{y}_i; \boldsymbol{P}))$, where $\boldsymbol{P}$ is a set of learnable parameters.

\subsubsection{Label Filter}
To filter noisy labels via our discovered distribution consistency pattern, we first randomly initialize several category centers $\boldsymbol{C} \in \mathcal{R}^{m \times k}$, i.e., $\boldsymbol{C} = \{\boldsymbol{c}_j\}_{j=1}^m$, where $m$ is the number of categories and $k$ is the length of hash code, ensuring that each category occupies a distinct and non-overlapping region. Then for each instance's hash code $\boldsymbol{\hat{b}}_i$, we calculate its similarity scores $\boldsymbol{d}_{ij}$ with each category center $\boldsymbol{c}_j$, denoted as:
\begin{equation}
    \label{similarity scores}
    \begin{aligned}
        \boldsymbol{d}_{ij} = \frac{\boldsymbol{\hat{b}}_i}{\lVert \boldsymbol{\hat{b}}_i \rVert} (\frac{\boldsymbol{c}_j}{\lVert \boldsymbol{c}_j \rVert})^T
    \end{aligned}
\end{equation}
By aggregating the similarity scores, we obtain the similarity scores matrix $\boldsymbol{D} \in \mathcal{R}^{n \times m}$ which captures the similarity of hash code to each category center. In this matrix, when a hash code $\boldsymbol{\hat{b}}_i$ belongs to a particular category $\boldsymbol{c}_j$, the corresponding similarity score $\boldsymbol{d}_{ij}$ tends to be high; otherwise, it is relatively low. Next, based on the consistency pattern between the 1-0 distribution of labels and the high-low distribution of similarity scores, we calculate the consistency level $\boldsymbol{T} = \{\boldsymbol{t}_i\}_{i=1}^n$ between the label distribution $\boldsymbol{l}_i$ and similarity distribution $\boldsymbol{d}_i$. We then design a consistency-based criterion to filter the dataset $\boldsymbol{O}$ into the clean label set $\boldsymbol{O_c}$ and noisy label set $\boldsymbol{O_n}$, which can be formulated as:
\begin{equation}
    \boldsymbol{t}_i = \frac{\sum_{j=1}^m \boldsymbol{l}_{ij} \boldsymbol{d}_{ij}}{\sum_{j=1}^m \boldsymbol{l}_{ij}}
\end{equation}
\begin{equation}
    \label{Oc}
    \begin{aligned}
        \boldsymbol{O_c} = \{(\boldsymbol{x}_i, \boldsymbol{y}_i, \boldsymbol{l}_i)|\boldsymbol{t}_i > \boldsymbol{\epsilon(\tau)}\}
    \end{aligned}
\end{equation}
\begin{equation}
    \label{On}
    \begin{aligned}
        \boldsymbol{O_n} = \{(\boldsymbol{x}_i, \boldsymbol{y}_i, \boldsymbol{l}_i)|\boldsymbol{t}_i \leq \boldsymbol{\epsilon(\tau)}\}
    \end{aligned}
\end{equation}
where $\boldsymbol{t}_i$ is employed to measure the degree of consistency between the label distribution and similarity distribution, with higher values indicating greater consistency, $\tau$ represents the noise ratio and $\boldsymbol{\epsilon(\tau)}$ denotes the filtering threshold to ensure $\tau{n}$ instances are identified as noisy label instances.

\subsubsection{Label Reconstructor}
To accurately learn the semantic knowledge and associations between labels and hash codes in the subsequent steps, we reconstruct the labels in $\boldsymbol{O_c}$ and $\boldsymbol{O_n}$. On the one hand, since the labels in $\boldsymbol{O_c}$ are clean, we directly use the initial labels as the reconstructed labels and adopt a standard measure in multi-modal hashing to establish semantic associations between hash codes and labels to learn implicit knowledge. On the other hand, for instances in $\boldsymbol{O_n}$, recognizing the significant advantage of labels as guiding information, we design a corrector via distribution consistency pattern to correct high-confidence noisy labels. Specifically, we treat the clean label set $\boldsymbol{O_c}$ as a knowledge base, and for a given instance $\boldsymbol{o}_i = \{\boldsymbol{x}_i, \boldsymbol{y}_i, \boldsymbol{l}_i\}$ in the noisy label set $\boldsymbol{O_n}$, we identify the two instances $\boldsymbol{o}_j = \{\boldsymbol{x}_j, \boldsymbol{y}_j, \boldsymbol{l}_j\}$ and $\boldsymbol{o}_k = \{\boldsymbol{x}_k, \boldsymbol{y}_k, \boldsymbol{l}_k\}$ from $\boldsymbol{O_c}$ whose distribution of similarity scores $\boldsymbol{d}_j$ and $\boldsymbol{d}_k$ most closely match that of $\boldsymbol{o}_i$. Here if the labels $\boldsymbol{l}_j$ and $\boldsymbol{l}_k$ of $\boldsymbol{o}_j$ and $\boldsymbol{o}_k$ are consistent, we infer that $\boldsymbol{o}_i$ has a high confidence of sharing the same label; otherwise, we consider $\boldsymbol{o}_i$ as having low confidence and treat it as unlabeled data. This corrector can be represented as follows:
\begin{equation}
    \label{noisy-clean-sim}
    \begin{aligned}
        \boldsymbol{m}_i = \boldsymbol{d}_i (\boldsymbol{D_c})^T
    \end{aligned}
\end{equation}
\begin{equation}
    \label{corrector}
    \begin{aligned}
        \{\boldsymbol{o}_j, \boldsymbol{o}_k\} = \arg\min_{\boldsymbol{o}_j, \boldsymbol{o}_k \in \boldsymbol{O_c}} \left( \text{rank}(\boldsymbol{m}_i, j) + \text{rank}(\boldsymbol{m}_i, k) \right)
    \end{aligned}
\end{equation}
\begin{equation}
    \boldsymbol{l}_i = \left \{ 
    \begin{aligned}
        &\boldsymbol{l}_j,  &\boldsymbol{l}_j = \boldsymbol{l}_k; \\
        &None, &otherwise.
    \end{aligned}
    \right.
\end{equation}
where $\boldsymbol{D_c}$ is the similarity scores matrix of clean label set $\boldsymbol{O_c}$, $\boldsymbol{m}_i$ represents the level of consistency between the similarity distribution of the noisy label instance $\boldsymbol{o}_i$ and each instance in the clean label set $\boldsymbol{O}_c$, and $\text{rank}(\cdot,\cdot)$ indicates the ranking of consistency levels. As for low-confidence noisy labels, we discard the original labels and treat them as unlabeled. Consequently, the noisy label set $\boldsymbol{O_n}$ is further divided into a corrected label set $\boldsymbol{O_r}$ and an unlabeled set $\boldsymbol{O_u}$.

\subsection{Objective Function and Optimization}
\subsubsection{Pointwise Learning for Clean Label Set}
To ensure that the generated fused hash codes accurately reflect label semantics, inspired by STBMH \cite{tu2024similarity}, we design the following pointwise loss on the clean label set $\boldsymbol{O_c}$ to make the hash codes as close as possible to their corresponding categories while keeping them distant from non-relevant categories, denoted as:
\begin{equation}
    \label{loss_o}
    \begin{aligned}
		\mathcal{L}_{o} =- \frac{1}{n_c} \sum\limits_{i}^{n_c}\sum\limits_{j\in\mathcal{Y}_i}log\frac{exp(\frac{1}{k}\boldsymbol{\hat{b}}_i^T\boldsymbol{c}_j)}{exp(\frac{1}{k}\boldsymbol{\hat{b}}_i^T\boldsymbol{c}_j) + \sum\limits_{h\in\mathcal{N}_i}exp(\frac{1}{k}\boldsymbol{\hat{b}}_i^T\boldsymbol{c}_h)},
    \end{aligned}
\end{equation}
where $k$ is the length of hash code, $\boldsymbol{n}_c$ is the number of instances in $\boldsymbol{O_c}$, $\mathcal{Y}_i$ contains the indices of all categories to which instance $\boldsymbol{o}_i$ belongs and $\mathcal{N}_i$ contains the indices of categories to which it does not belong. By minimizing $\mathcal{L}_{o}$, the value of $exp(\frac{1}{k}\boldsymbol{\hat{b}}_i^T\boldsymbol{c}_j)$ becomes significantly higher than that of $\sum\limits_{h\in\mathcal{N}_i}exp(\frac{1}{k}\boldsymbol{\hat{b}}_i^T\boldsymbol{c}_h)$, indicating that the similarity between the hash code and its corresponding category center is high, while the similarity with non-relevant category centers is low, thereby achieving the desired objective. 

\subsubsection{Pairwise Learning for Corrector Label Set}
Meanwhile,  for the corrected label set $\boldsymbol{O_r}$, since the labels within still have the potential for error correction, we adopt a pairwise loss to emphasize the relative similarity relationships among instances as a whole, rather than direct associations between instances and specific individual categories, expressed as follows:
\begin{equation}
    \label{loss_a}
    \begin{aligned}
		\mathcal{L}_{a} = \sum\limits_i^{n_r} \sum\limits_j^{n_r} \|cos(\boldsymbol{\hat{b}}_i, \boldsymbol{\hat{b}}_j) - \boldsymbol{s}_{ij}\|^2_F
    \end{aligned}
\end{equation}
\begin{equation}
    \label{cos}
    \begin{aligned}
		cos(\boldsymbol{\hat{b}}_i, \boldsymbol{\hat{b}}_j)= \frac{\boldsymbol{\hat{b}}_i^T\boldsymbol{\hat{b}}_j}{\|\boldsymbol{\hat{b}}_i\|\|\boldsymbol{\hat{b}}_j\|}=\frac{1}{k}\boldsymbol{\hat{b}}_i^T\boldsymbol{\hat{b}}_j
    \end{aligned}
\end{equation}
where $k$ is the length of hash code, $\boldsymbol{n}_r$ is the number of instances in $\boldsymbol{O_r}$, $cos(\cdot,\cdot)$ is employed to measure the cosine similarity between hash codes and $\boldsymbol{s}_{ij}$ represents the pairwise similarity defined by the labels. By minimizing $\mathcal{L}_{a}$, the cosine similarity between hash codes of similar instances defined by the labels will approach 1 and their Hamming distance $d_H(\boldsymbol{\hat{b}}_i, \boldsymbol{\hat{b}}_j)$ will decrease, where $d_H(\boldsymbol{\hat{b}}_i, \boldsymbol{\hat{b}}_j)=\frac{1}{2}(k - \boldsymbol{\hat{b}}_i^T\boldsymbol{\hat{b}}_j)$.

\subsubsection{Unsupervised Learning for Unlabeled Set}
For the unlabeled set $\boldsymbol{O_u}$, given that text and image data points inherently contain rich semantic knowledge, we employ unsupervised contrastive learning to uncover hidden semantic relationships and further enhance model performance. Specifically, for an instance $\boldsymbol{o}_i$ in $\boldsymbol{O_u}$, we first generate an augmented instance $\boldsymbol{o}_i^{'}$ through data augmentation. Then due to $\boldsymbol{o}_i$ and $\boldsymbol{o}_i^{'}$ embody the same semantic meaning, their corresponding hash codes should be as consistent as possible. Therefore, we design the following contrastive loss to minimize the distance between hash code $\boldsymbol{\hat{b}}_i$ and $\boldsymbol{\hat{b}}_i^{'}$:
\begin{equation}
    \label{loss_u}
    \begin{aligned}
        \mathcal{L}_{u} = \frac{1}{n_u} \sum_{i=1}^{n_u} \left(1 - s_{ii}\right) + \frac{1}{{n_u}^2} \sum_{i=1}^{n_u} \sum_{j=1, j \neq i}^{n_u} \max(0, s_{ij} - \epsilon)
    \end{aligned}
\end{equation}
where $s_{ij} = cos(\boldsymbol{\hat{b}}_i, \boldsymbol{\hat{b}}_j^{'})$ i.e. cosine similarity of hash codes, $\boldsymbol{n}_u$ is the number of instances in $\boldsymbol{O_u}$ and $\epsilon$ is the threshold that restricts the similarity between different instance pairs. By minimizing $\mathcal{L}_{u}$, the similarity between the same instance pairs will approach 1, while the similarity between different instance pairs will be less than the threshold $\epsilon$, thus ensuring that more similar instances are mapped to closer hash codes and achieving the goal of extracting the inherent semantics of the instance through unsupervised learning.

\subsubsection{Center Learning and Quantization}
In addition, each category should occupy a distinct region without interference. To achieve this, the centers of the categories must be well-separated, which lead us to design the following center loss:
\begin{equation}
    \label{dij}
    \begin{aligned}
        d_{ij} = |\boldsymbol{c}_i - \boldsymbol{c}_j\|^2_F
    \end{aligned}
\end{equation}
\begin{equation}
    \label{loss_c}
    \begin{aligned}
        \mathcal{L}_{c} = -\frac{1}{\left|\{d_{ij} | i < j\}\right|} \sum_{i < j} d_{ij} - \min_{i < j} d_{ij}
    \end{aligned}
\end{equation}
where $d_{ij}$ represents the distance between two category centers and $i < j$ indicates that only the upper triangular region is considered. By minimizing $\mathcal{L}_{c}$, sufficient separation between different categories is achieved which enhances the model's ability to distinguish them. Furthermore, as the model ultimately relies on the $sgn(\cdot)$ function to convert the fused hash codes into binary hash codes, we introduce the following loss to control the quantization loss incurred during this process:
\begin{equation}
    \label{loss_q}
    \begin{aligned}
        \mathcal{L}_{q} = \sum\limits_i^{n}\|\boldsymbol{\hat{b}}_i - \boldsymbol{b}_i\|^2_F
    \end{aligned}
\end{equation}

\subsubsection{Overall Objective Function}
Finally, by combining the losses from the above components, the overall objective function of the multi-modal hashing network is defined as:
\begin{equation}
    \label{loss}
    \begin{aligned}
        \mathcal{L} = \mathcal{L}_{o} + \alpha \mathcal{L}_{a} + \beta \mathcal{L}_{u} + \gamma \mathcal{L}_{c} + \eta \mathcal{L}_{q}
    \end{aligned}
\end{equation}
where $\alpha$, $\beta$, $\gamma$ and $\eta$ are hyperparameters. By minimizing this loss during the training of the whole multi-modal hashing network, the model can generate high-quality and distinguishable hash codes even in the presence of noisy labels, which significantly enhances its performance. The detailed algorithm is listed in the Algorithm \ref{alg}.

\begin{algorithm}[]
    \caption{Learning algorithm for DCGMH}
    \label{alg}
    \begin{algorithmic}[1]
		\REQUIRE
		Instances $\boldsymbol{O}=\{\boldsymbol{o}_i\}_{i=1}^N$, each instances $\boldsymbol{o}_i$ contains an image $\boldsymbol{x}_i$, a text $\boldsymbol{y}_i$ and label $\boldsymbol{l}_i$, the length of hash codes $k$, and the hyper-parameter $\alpha$, $\beta$, $\gamma$ and $\eta$.
		\ENSURE 
		Multi-model hashing network.
		\STATE Initialize the set of parameters $\boldsymbol{P}$ of the multi-model hashing network, the learning rate, the category centers $C$ and mini-batch size  $z=48$.
		\REPEAT
		\FOR{$i=1:\frac{N}{z}$}
        \STATE Randomly sample $z$ instances from the training set as a mini-batch.
        \STATE Generate the hash code $\boldsymbol{\hat{b}}_i$ with $\boldsymbol{o}_i$ as input.
        \STATE Filter out clean label set $\boldsymbol{O_c}$ and noisy label set $\boldsymbol{O_n}$ based on the distribution consistency pattern.
        \STATE Correct high-confidence noisy label in $\boldsymbol{O_n}$ and divide $\boldsymbol{O_n}$ into a corrected label set $\boldsymbol{O_r}$ and an unlabeled set $\boldsymbol{O_u}$.
        \STATE Independently learn the semantic information and relationships for $\boldsymbol{O_c}$, $\boldsymbol{O_r}$ and $\boldsymbol{O_u}$, and then aggregate them.
        \STATE Update the parameters of the hashing model by minimizing $\mathcal{L}$ through the backpropagation algorithm.
		\ENDFOR
		\UNTIL{Convergence}
    \end{algorithmic}
\end{algorithm}

\begin{table*}[]
    \centering
    \begin{tabular}{c|cccc|cccc|cccc}
        \toprule[1.2pt]
        \multirow{2}{*}{Method} & \multicolumn{4}{c|}{MIR Flickr}      & \multicolumn{4}{c|}{NUS-WIDE}  & \multicolumn{4}{c}{MS COCO}    \\ \cline{2-13} 
        &16 &32 &64 &128 &16 &32 &64 &128 &16 &32 &64 &128  \\ \hline
        FOMH    &0.697 &0.726	&0.733	&0.742	&0.586	&0.604	&0.657	&0.662	&0.425 &0.440 &0.468 &0.485 \\
        OMHDQ   &0.762 &0.771	&0.777	&0.783	&0.672	&0.689	&0.704	&0.717	&0.522 &0.538 &0.546 &0.573 \\
        SDMH    &0.772 &0.782	&0.794	&0.798	&0.666	&0.690	&0.698	&0.713	&0.396 &0.398 &0.399 &0.399 \\
        SAPMH   &0.751 &0.783 &0.800 &0.806	&0.639	&0.670	&0.677	&0.692	&0.438 &0.442 &0.458 &0.463 \\
        DCMVH   &0.715 &0.725	&0.742	&0.754	&0.590	&0.612	&0.667	&0.685	&0.399 &0.414 &0.477 &0.499 \\
        BSTH    &0.705 &0.715	&0.736	&0.757	&0.559 &0.591	&0.615	&0.640	&0.503 &0.524 &0.538 &0.561 \\
        NCH     &0.751 &0.759	&0.769	&0.778	&0.618	&0.634	&0.641	&0.664	&0.517 &0.533 &0.551 &0.569 \\
        GCIMH   &0.738 &0.746	&0.751	&0.756	&0.600	&0.604	&0.618	&0.623	&0.488 &0.499 &0.516 &0.523 \\
        STBMH   &0.748 &0.767	&0.779	&0.795	&0.622	&0.656	&0.681	&0.691	&0.509 &0.540 &0.564 &0.589 \\
        DIOR    &0.750 &0.781	&0.820	&0.835	&0.636	&0.692	&0.737	&0.754	&0.482 &0.509 &0.560 &0.596 \\
        \midrule
        Ours    &\textbf{0.796}	&\textbf{0.823}	&\textbf{0.846}	&\textbf{0.850}	&\textbf{0.717}	&\textbf{0.739}	&\textbf{0.755}	&\textbf{0.757}	&\textbf{0.551}	&\textbf{0.591}    &\textbf{0.631}    &\textbf{0.649} \\ \bottomrule
    \end{tabular}
    \caption{MAPs at a noisy label ratio of 40\% for different hash code lengths across the three benchmark datasets.}
    \label{map}
\end{table*}

\subsection{Out-of-Sample Extension}
After optimizing the multi-modal hashing network by minimizing $\mathcal{L}$, we can use the model to generate hash codes for instances outside the training set. Specifically, for a given unseen instance $\boldsymbol{o}_i = \{\boldsymbol{x}_i, \boldsymbol{y}_i, \boldsymbol{l}_i\}$, we obtain the binary hash code by feeding the instance into the optimized multi-modal hashing network, denoted as:
\begin{equation}
    \begin{aligned}
        \boldsymbol{b}_i = sgn(\mathcal{F}(\boldsymbol{x}_i, \boldsymbol{y}_i; \boldsymbol{P}))
    \end{aligned}
\end{equation}

\section{Experiments}
\subsection{Experimental Setting}

\begin{figure*}[htbp]
	\centering
        \subfigure[MIR Flickr]{
            \includegraphics[width=0.3\textwidth]{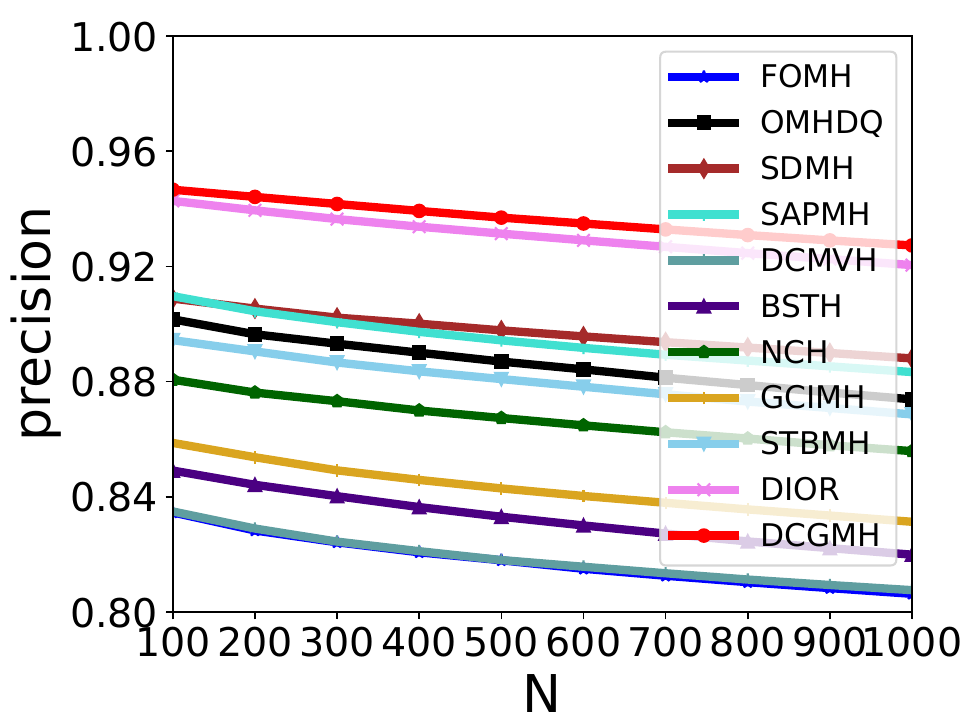}
        }
        \subfigure[NUS-WIDE]{
            \includegraphics[width=0.3\textwidth]{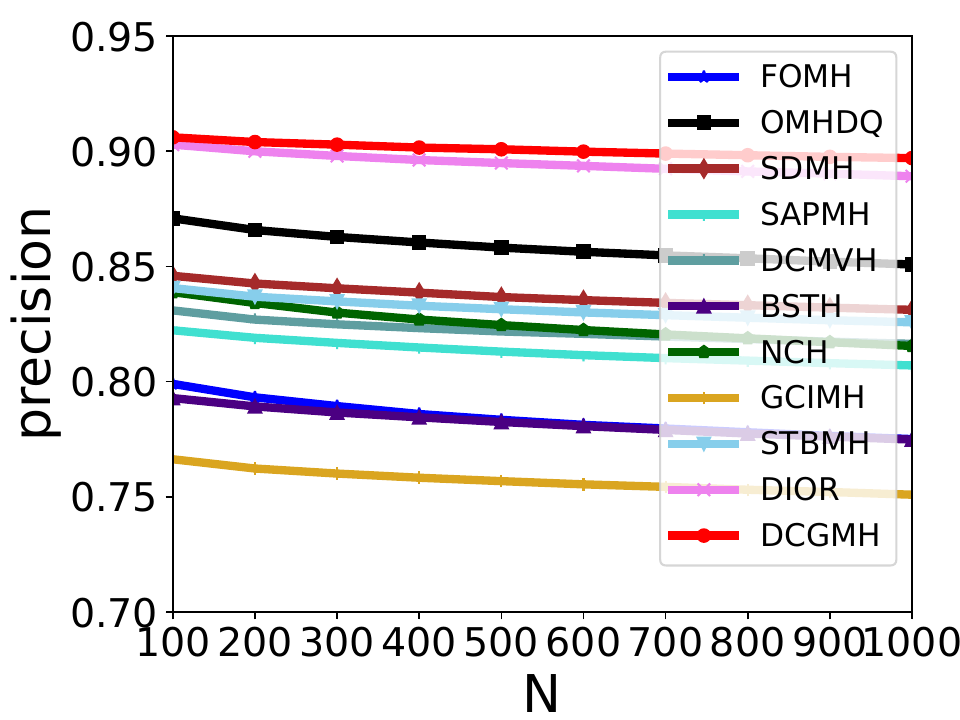}
        }
        \subfigure[MS COCO]{
            \includegraphics[width=0.3\textwidth]{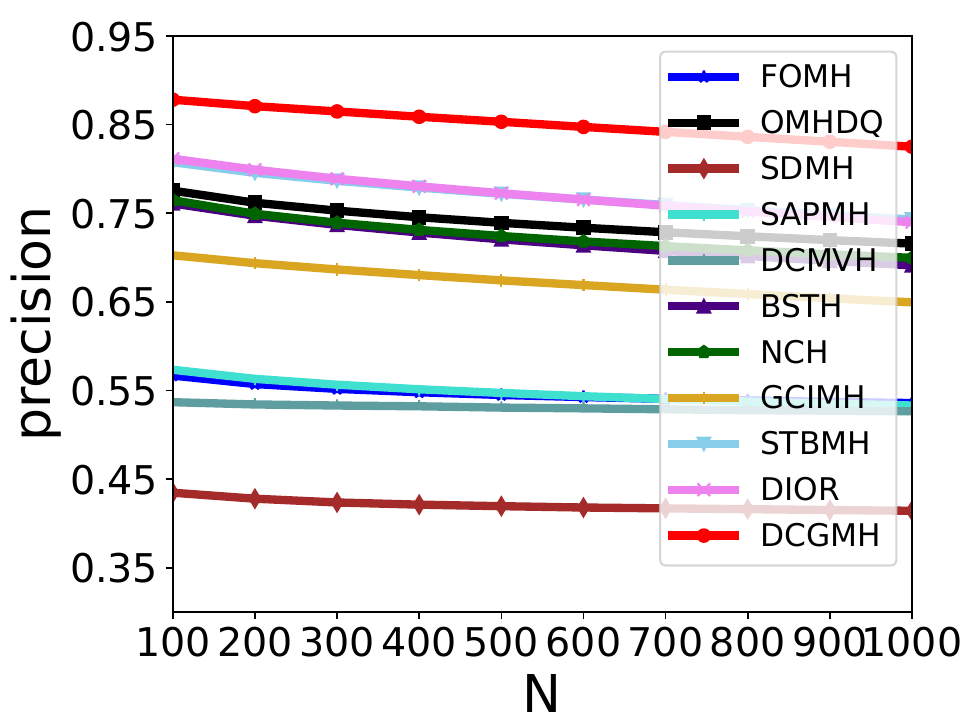}
        }
	\caption{\centering Precision@N curves at a noisy label ratio of 40\% on the three benchmark datasets with the code length of 64 bits.}
	\label{pn}
\end{figure*} 

\begin{figure*}[]
	\centering
        \subfigure[MIR Flickr]{
            \includegraphics[width=0.3\textwidth]{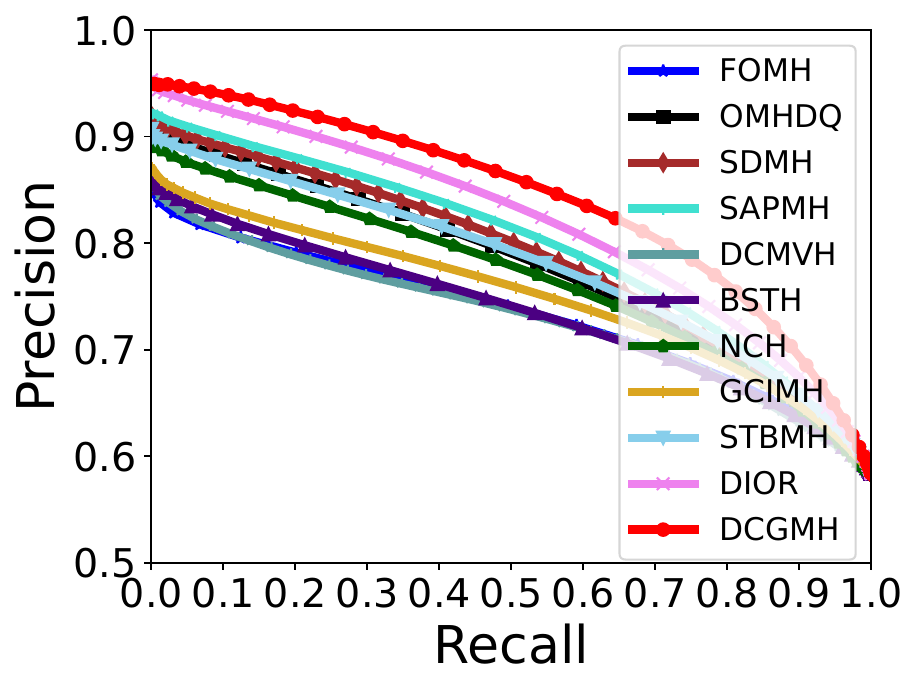}
        }
        \subfigure[NUS-WIDE]{
            \includegraphics[width=0.3\textwidth]{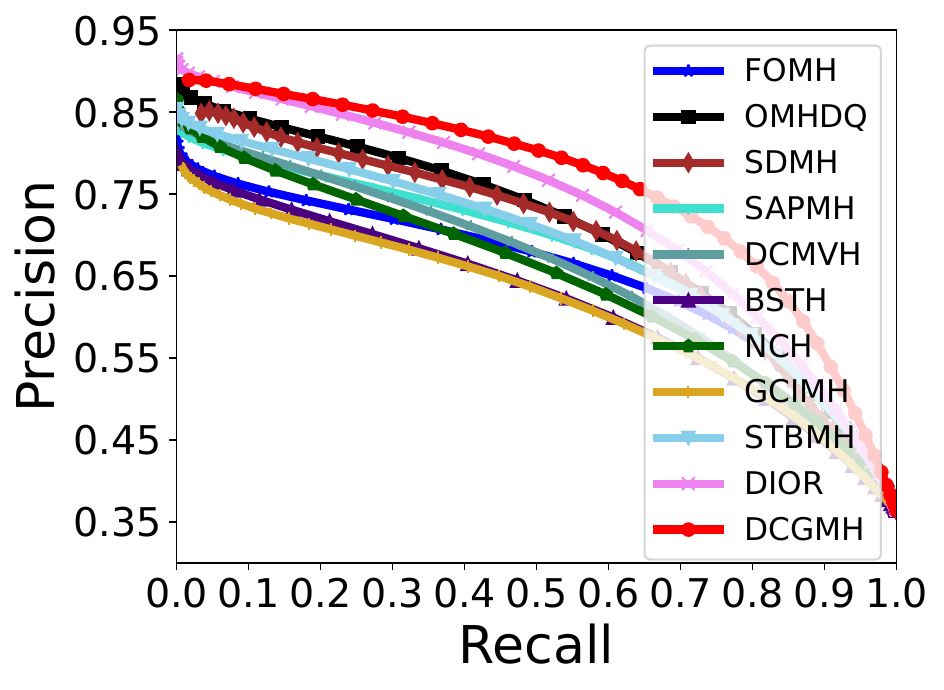}
        }
        \subfigure[MS COCO]{
            \includegraphics[width=0.3\textwidth]{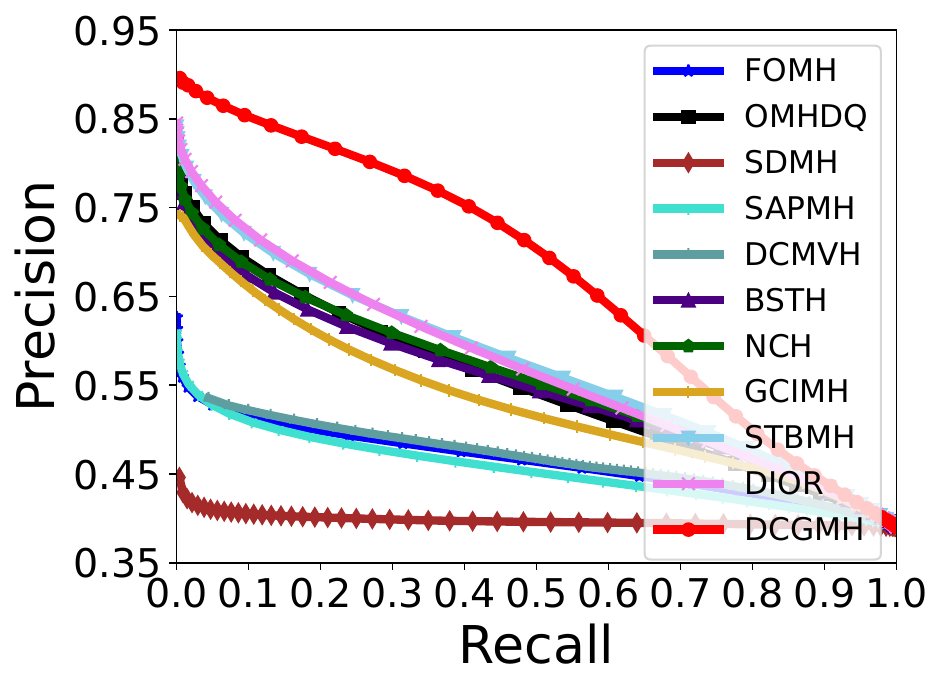}
        }
	\caption{\centering PR curves at a noisy label ratio of 40\% on the three benchmark datasets with the code length of 64 bits.}
	\label{pr}
\end{figure*} 

\subsubsection{Datasets}
To validate the effectiveness of our proposed method, we conduct experiments on three widely used image-text benchmark datasets: MIR Flickr \cite{huiskes2010new}, NUS-WIDE \cite{chua2009nus}, and MS COCO \cite{lin2014microsoft}. Among the datasets, the MIR Flickr dataset, which includes 24 labels, contains 20,015 data pairs, with each text represented as a 1,386-dimensional BoW vector. We partition this dataset into a testing set of 2,243 pairs, a retrieval set of 17,772 pairs, and a training set of 5,000 pairs. The NUS-WIDE dataset, featuring 21 common labels, includes 195,834 pairs with text represented by 1,000-dimensional BoW vectors. We derive a testing set of 2,085 pairs, a retrieval set of 193,749 pairs, and a training set of 21,000 pairs. For MS COCO, which includes 80 labels, we select 5,981 pairs for testing, 82,783 pairs for retrieval, and 18,000 pairs for training.

Given that the labels in these datasets are originally correctly annotated, we introduce noise to better simulate the performance of the proposed method in real-world scenarios with noisy labels. Specifically, when setting the noisy label ratio to $w$, we randomly select $w$ proportion of training instances as noisy label instances and averagely categorize them into four types to accurately simulate real-world scenarios: (1) maintaining the number of categories that belong and including some original categories; (2) maintaining the number of categories that belong but excluding the original categories; (3) changing the number of categories that belong while including some original categories; and (4) changing the number of categories that belong and excluding the original categories.

\subsubsection{Baselines}
To evaluate the superiority of our proposed method, we compare it with ten state-of-the-art multi-modal hashing methods, i.e., FOMH \cite{lu2019flexible}, OMHDQ \cite{lu2019online}, SDMH \cite{lu2019efficient}, SAPMH \cite{zheng2020adaptive}, DCMVH \cite{zhu2020deep}, BSTH \cite{tan2022bit}, NCH \cite{tan2023partial}, GCIMH \cite{shen2023graph}, STBMH \cite{tu2024similarity} and DIOR \cite{wang2023dior}, where the first four adopt shallow frameworks and the latter six employ deep networks. Notably, DIOR is originally developed for noisy label handling in image hashing; here, we adapt it for multi-modal hashing while retaining its core structure. 

\subsubsection{Implementation Details}
Similar to DIOR, we first perform warm-up training without noisy label filtering and correction to allow the model to learn the basic hashing mapping capabilities, with the warm-up epochs set to 5, 5, and 30 for the MIR Flickr, NUS-WIDE, and MS COCO respectively. During the training of the hashing network on a single NVIDIA RTX 3090Ti GPU, the SGD optimizer with a batch size of 48 is adopted for parameter optimization, with an initial learning rate set to 0.005, 0.001, and 0.01 for the three datasets, respectively. Both the image and text modalities employ MLP architectures consisting of two linear projection layers ($d_m$ $\rightarrow$ 4,096 $\rightarrow$ 128$k$), where $d_m$ is the feature dimension of the modality data points and $k$ is the length of the hash codes. Regarding hyper-parameters, $\alpha$ and $\beta$ are set to 1, 1.5, 1.2 and 0.15, 0.05, 0.2 for the three datasets, respectively. $\gamma$ and $\eta$ are empirically set to 5 and 1 for all three datasets, respectively, which will be discussed later. Additionally, all experimental results are averaged over three runs with different random seeds.

\subsubsection{Evaluation Protocol}
In our experiment, we evaluate the proposed model's efficiency and effectiveness using two protocols: Hamming ranking and hash lookup. For the Hamming ranking protocol, which relies on Hamming distances, we use Mean Average Precision (MAP) and Precision curves at different top N results (P@N) to measure accuracy. Meanwhile, the hash lookup protocol's accuracy is assessed using the Precision-Recall (PR) curve within a specified Hamming radius. Furthermore, experiments are conducted across two sub-tasks: performance comparison at a fixed noisy label ratio, and performance variation with different noisy label ratios.

\subsection{Experimental Results}
\subsubsection{Performance Comparison at a Fixed Noisy Label Ratio}

To simulate the model's retrieval performance in real-world noisy label scenarios, we conduct experiments on three datasets assuming 40\% noisy labels in the training set, with the MAP, P@N and PR curve comparison with all baselines shown in Table \ref{map}, Figure \ref{pn} and Figure \ref{pr}. 

The following conclusions can be drawn from the experimental results: (1) In the scenario with 40\% noisy labels, our proposed method DCGMH achieves the best MAP performance compared to all baselines. Specifically, when compared to existing multi-modal hashing models that do not account for noisy labels, DCGMH provides average improvements of at least 4.9\%, 6.7\%, and 9.3\% on the MIR Flickr, NUS-WIDE, and MS COCO, respectively. Additionally, DCGMH significantly outperforms the DIOR model designed to handle noisy labels. This demonstrates DCGMH's effectiveness in generating high-quality hash codes in noisy label scenarios. (2) DCGMH consistently obtains the highest P@N scores, which suggests that DCGMH excels at retrieving the most relevant results more quickly and accurately when in scenarios with noisy labels. The superior P@N performance further emphasizes the model's robustness and effectiveness in mitigating the adverse effects of noisy labels, enhancing the multi-modal hashing model's retrieval performance. (3) The PR curve of DCGMH outperforms all baselines, showing that the hash codes generated by the model maintain smaller Hamming distances for more similar instances, thereby maximizing the preservation of the instance's intrinsic semantic information and relationships. (4) As the length of the hash codes increases, both DCGMH and the baseline methods show improved performance. This indicates that longer hash codes enable a more detailed analysis of instance structure and features, allowing for better preservation of semantic information and consequently enhancing retrieval performance.

\subsubsection{Performance Variation with Different Noisy Label Ratios}
\begin{figure*}[]
	\centering
        \subfigure[MIR Flickr]{
            \includegraphics[width=0.3\textwidth]{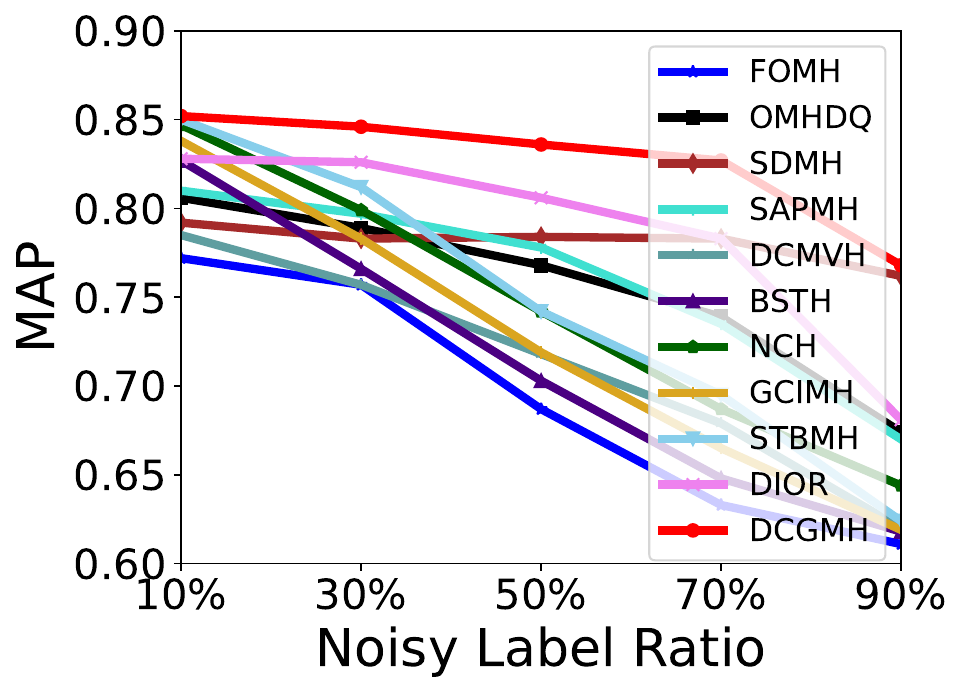}
        }
        \subfigure[NUS-WIDE]{
            \includegraphics[width=0.3\textwidth]{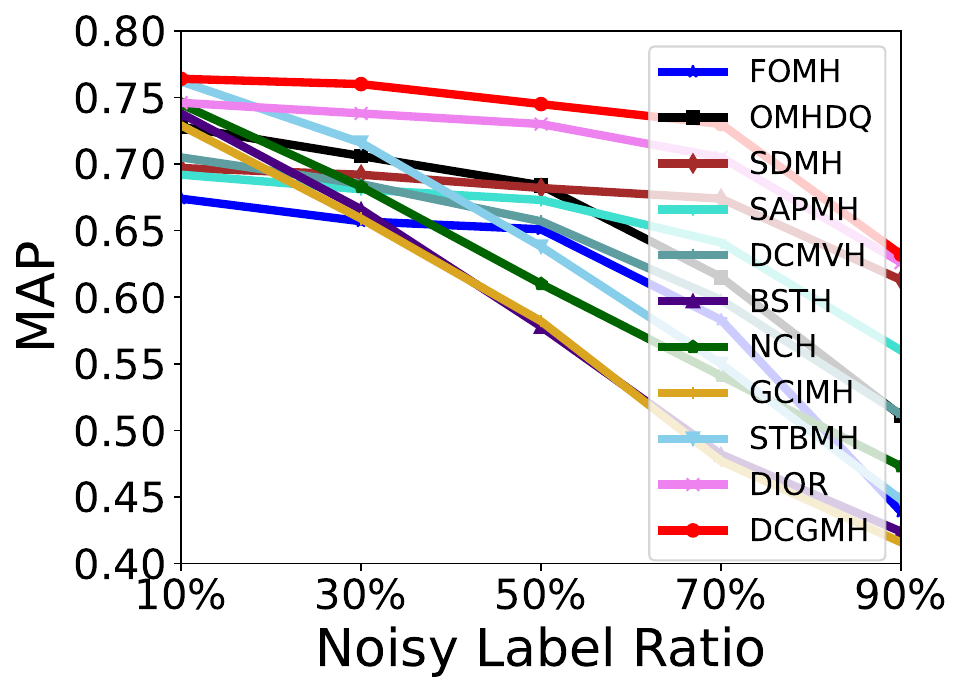}
        }
        \subfigure[MS COCO]{
            \includegraphics[width=0.3\textwidth]{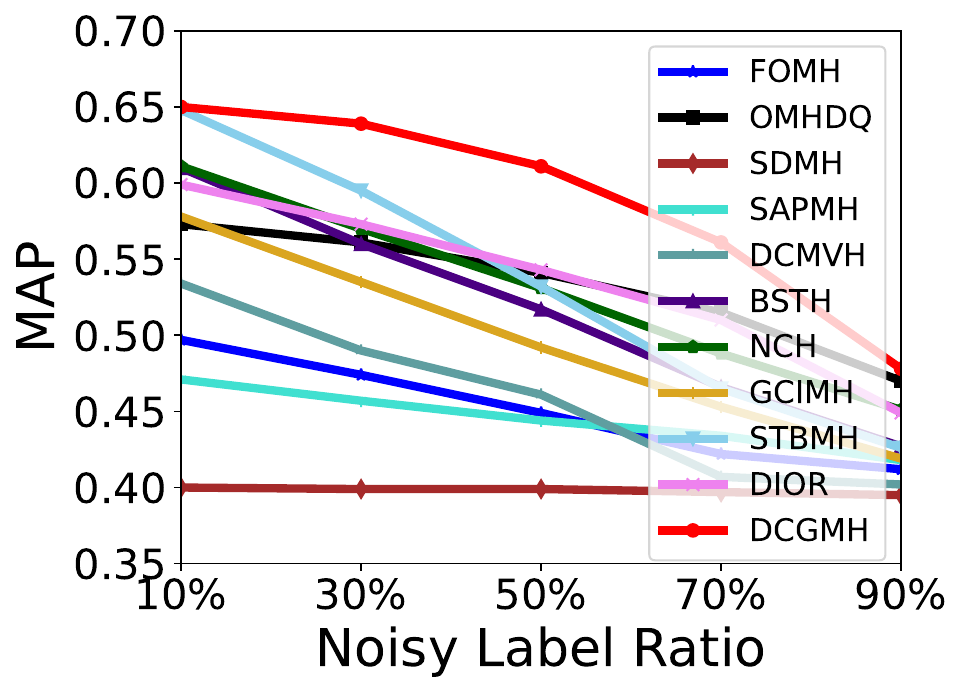}
        }
	\caption{\centering MAPs at different noisy label ratios on the training set of three benchmark datasets with the code length of 64 bits.}
	\label{ratio}
\end{figure*} 

\begin{table*}[]
    \centering
    \begin{tabular}{c|cccc|cccc|cccc}
        \toprule[1.2pt]
        \multirow{2}{*}{Method} & \multicolumn{4}{c|}{MIR Flickr}      & \multicolumn{4}{c|}{NUS-WIDE}  & \multicolumn{4}{c}{MS COCO}    \\ \cline{2-13} 
        &16 & 32 & 64 & 128 &16 & 32 & 64 & 128 &16 & 32 & 64 & 128  \\ \hline
        DCGMH-I &0.718 &0.763 &0.784 &0.809 &0.586 &0.706 &0.721 &0.724 &0.512 &0.561 &0.610 &0.607 \\
        DCGMH-R	&0.786 &0.814 &0.840 &0.843 &0.705 &0.725 &0.736 &0.740 &0.501 &0.590 &0.624 &0.647 \\
        DCGMH-U &0.791 &0.822 &0.840 &0.847 &0.709 &0.732 &0.747 &0.755 &0.470 &0.587 &0.613 &0.637 \\
        DCGMH-RU &0.781 &0.811 &0.836 &0.839 &0.699 &0.722 &0.723 &0.731 &0.477 &0.587 &0.611 &0.635 \\
        DCGMH-D &0.770 &0.805 &0.828 &0.827 &0.678 &0.693 &0.701 &0.710 &0.469 &0.583 &0.622 &0.638 \\
        \midrule
        Ours    &\textbf{0.796}	&\textbf{0.823}	&\textbf{0.846}	&\textbf{0.850} &\textbf{0.717}	&\textbf{0.739}	&\textbf{0.755}	&\textbf{0.757}	&\textbf{0.551}	&\textbf{0.591}    &\textbf{0.631}    &\textbf{0.649} \\ \bottomrule[1.2pt]
    \end{tabular}
    \caption{MAPs of the variants at a noisy label ratio of 40\% on the three benchmark datasets.}
    \label{ablation}
\end{table*}
To evaluate how different noisy label ratios affect model performance, we examine the MAP metric under 64-bit hash codes as the noisy label ratio varied within the range of [10\%, 30\%, 50\%, 70\%, 90\%]. The experimental results are depicted in Figure \ref{ratio}, from which we can obtain the following observations: (1) Our proposed method DCGMH consistently achieves the best retrieval performance across nearly all noisy label ratios, which demonstrates that DCGMH effectively mitigates the negative impact of noisy labels in the training set and showcases its exceptional robustness. (2) As the noisy label ratio increases, DCGMH shows the slowest performance decline compared to other models, indicating its lower sensitivity to noisy labels. Even with a high proportion of noisy labels, DCGMH can effectively filter and correct them to maintain superior performance. (3) Notably, when the noisy label ratio reaches 90\%, the model's performance drops sharply. This decline is likely due to the scarcity of clean label instances, which hinders the model’s ability to effectively learn basic hash mappings during the warm-up phase, ultimately leading to overfitting and indistinguishable hash codes.

\begin{figure*}[]
	\centering
        \subfigure[]{
            \includegraphics[width=0.22\textwidth]{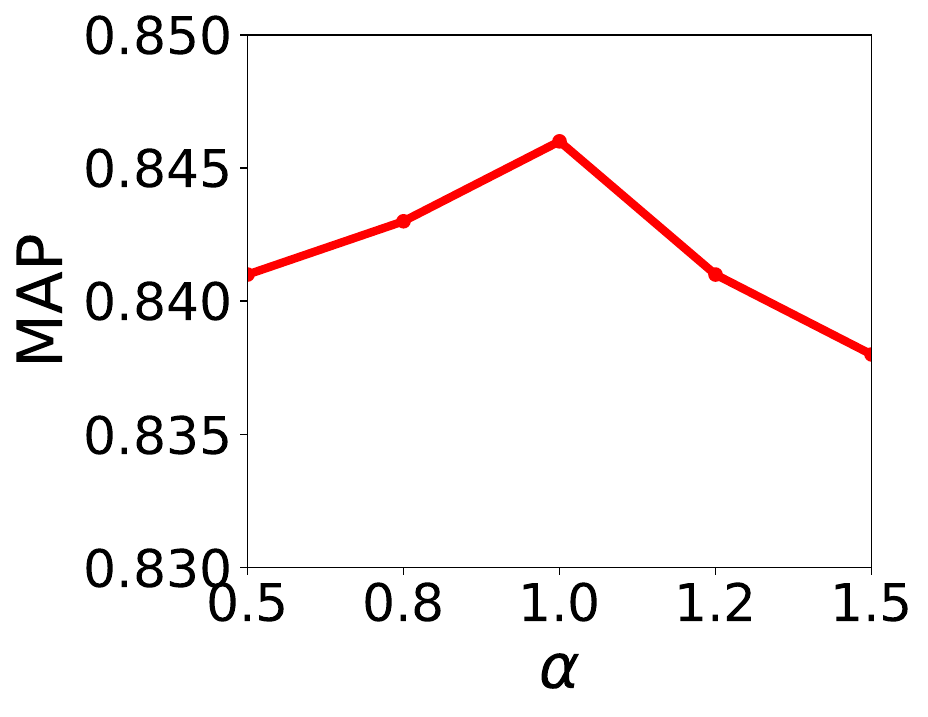}
        }
        \subfigure[]{
            \includegraphics[width=0.22\textwidth]{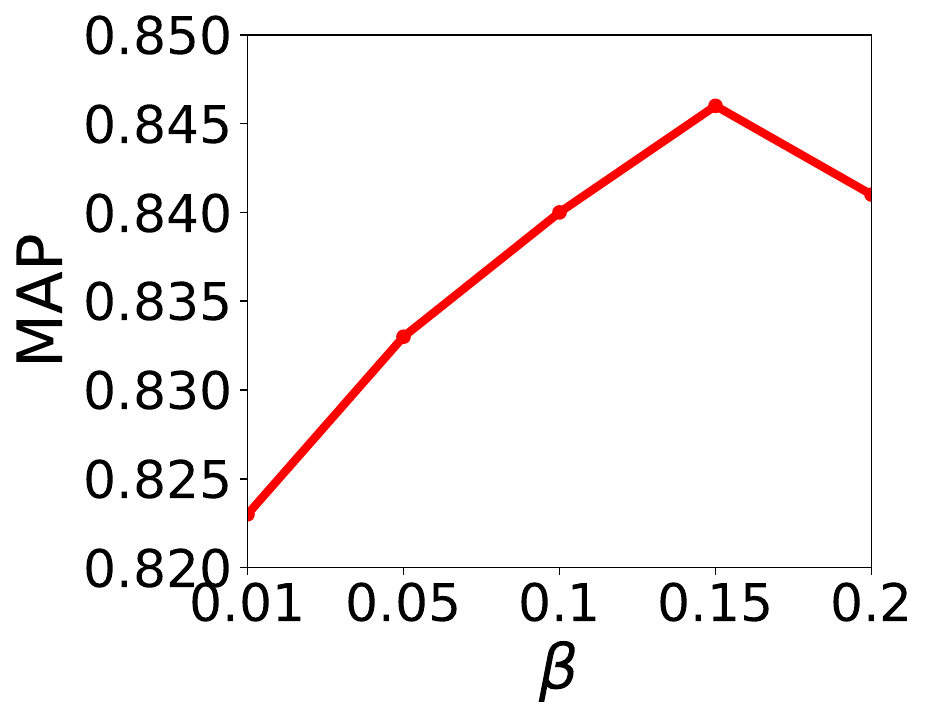}
        }
        \subfigure[]{
            \includegraphics[width=0.22\textwidth]{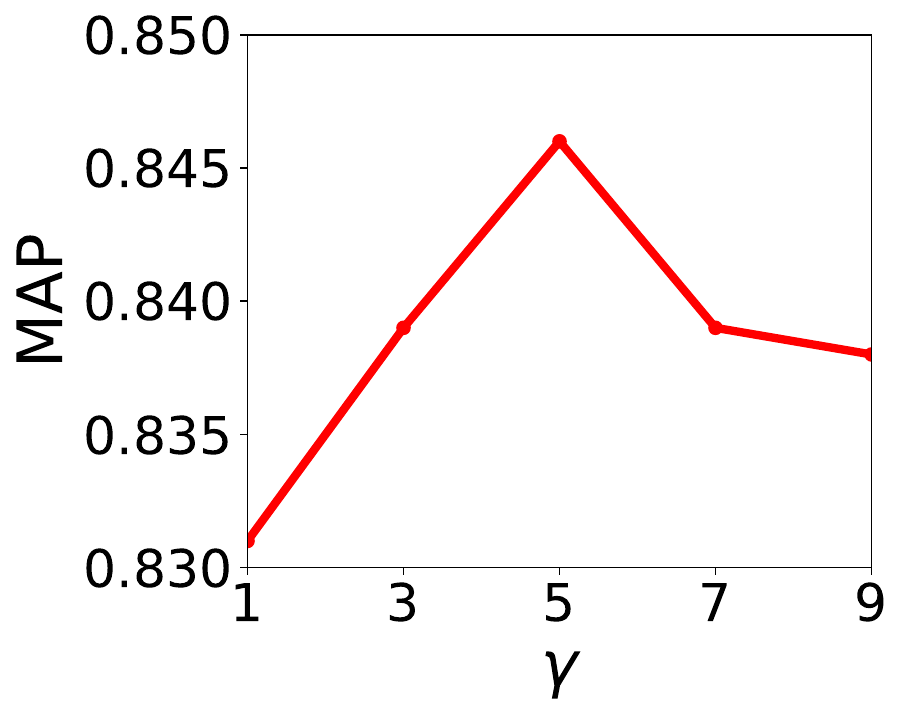}
        }
        \subfigure[]{
            \includegraphics[width=0.22\textwidth]{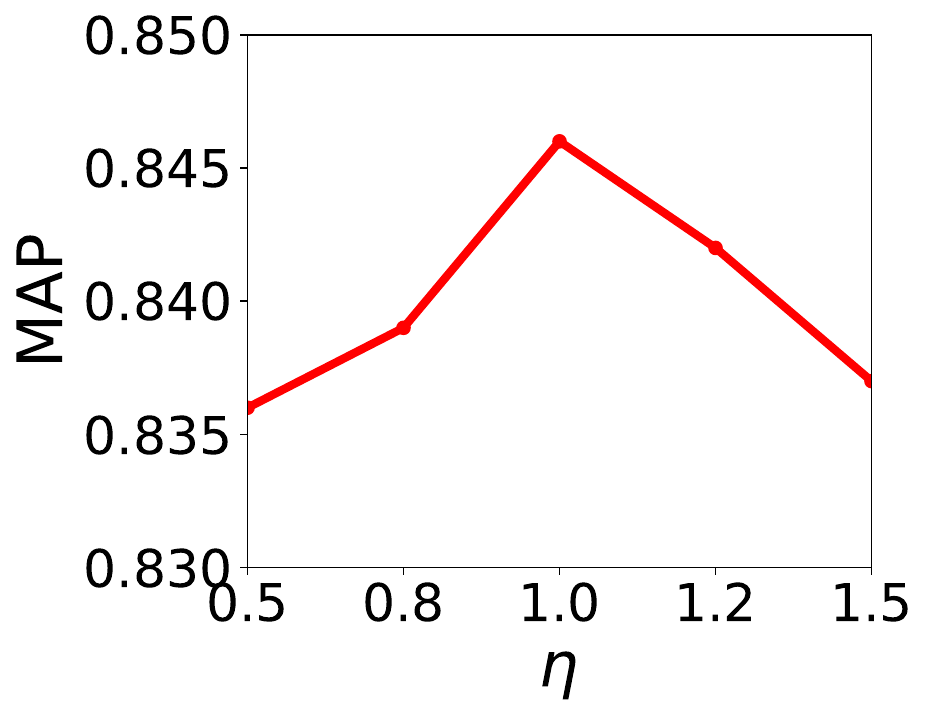}
        }
	\caption{Sensitivity analysis of hyper-parameters on MIR Flickr with 64-bit code length and 40\% noisy label ratio.}
	\label{hyper-param}
\end{figure*} 

\subsection{Ablation Study}
To evaluate the effectiveness of each module design and filtering strategy, we define the following five variants: DCGMH-I is trained directly on the original dataset without applying any noise label filtering or reconstruction; DCGMH-R treats all noisy labels as unlabeled for unsupervised learning, without correcting high-confidence noisy labels; DCGMH-U directly discards low-confidence noisy labels; DCGMH-RU trains solely on clean labels and discards all filtered noisy labels; DCGMH-D filter noisy labels via the difference in loss between different augmented views like DIOR. 

The experimental results are shown in Table \ref{ablation}, where the training set has 40\% noisy labels. From these results, we can draw the following conclusions: (1) Compared to DCGMH-I trained directly with noisy labels, both DCGMH and the other variants show substantial performance improvements, highlighting the critical role of the noisy label filtering module in preventing model overfitting and enhancing retrieval performance. (2) Among DCGMH-R, DCGMH-U, DCGMH-RU, and DCGMH, the latter achieves the best retrieval performance, indicating that both the noisy label correction module and unsupervised semantic learning module contribute to enhancing the hash model's ability to capture richer semantic information and generate more precise hash codes. (3) Compared to DCGMH-D, DCGMH delivers superior retrieval performance, demonstrating that our proposed filtering strategy more accurately and thoroughly filters out noisy labels to prevent model overfitting.

\subsection{Sensitivity to Hyper-parameters}
Taking the MIR Flickr dataset as an example, we explore the impact of different values of the hyperparameters $\alpha$, $\beta$, $\gamma$, and $\eta$ on model performance with the hash code length of 64 bits and the noisy label ratio of 40\%. The experimental results are shown in Figure \ref{hyper-param}, where the $\beta$ value varies between 0.01 and 0.2 based on empirical settings, while the $\alpha$ and $\eta$ values range from 0.5 to 1.5, and the $\gamma$ value ranges from 1 to 9. The results indicate that as the hyperparameter values increase, the model's performance initially improves and then declines. Moreover, regardless of the values selected within the range, the performance consistently surpasses that of existing state-of-the-art multi-modal hashing models. To achieve optimal performance, we set the values of $\alpha$, $\beta$, $\gamma$, and $\eta$ to 1, 0.15, 5, and 1, respectively, in the experiments. Similar trends are observed on the NUS-WIDE and COCO, where the corresponding hyperparameters are set to 1.5, 0.05, 5, 1, and 1.2, 0.2, 5, 1, respectively.

\section{Conclusion}
In this paper, we propose a novel Distribution-Consistency-Guided Multi-modal Hashing (DCGMH), which can filter out noisy labels via our discovered consistency pattern between the 1-0 distribution of labels and the high-low distribution of similarity scores and simultaneously design a corrector to correct high-confidence noisy labels to generate high-quality and discriminative hash codes, thereby further enhancing the model's retrieval performance in noisy label scenarios. Extensive experiments on three public datasets demonstrate that our proposed DCGMH outperforms the state-of-the-art baselines in multi-modal retrieval tasks.

\section{Acknowledgments}
The work is supported by National Natural Science Foundation of China (No. 62172039, 6240074151, U21B2009 and 62276110) and MIIT Program(CEIEC-2022-ZM02-0247).



\bibliography{aaai25}

\begin{thebibliography}{60}
\providecommand{\natexlab}[1]{#1}

\bibitem[{An et~al.(2022)An, Luo, Zhang, Zhu, and Lu}]{an2022cognitive}
An, J.; Luo, H.; Zhang, Z.; Zhu, L.; and Lu, G. 2022.
\newblock Cognitive multi-modal consistent hashing with flexible semantic transformation.
\newblock \emph{Information Processing \& Management}, 59(1): 102743.

\bibitem[{Chen et~al.(2020{\natexlab{a}})Chen, Zhang, Tian, Wang, Zhang, and Li}]{chen2020enhanced}
Chen, Y.; Zhang, H.; Tian, Z.; Wang, J.; Zhang, D.; and Li, X. 2020{\natexlab{a}}.
\newblock Enhanced discrete multi-modal hashing: More constraints yet less time to learn.
\newblock \emph{IEEE Transactions on Knowledge and Data Engineering}, 34(3): 1177--1190.

\bibitem[{Chen et~al.(2020{\natexlab{b}})Chen, Li, Yu, El~Kholy, Ahmed, Gan, Cheng, and Liu}]{chen2020uniter}
Chen, Y.-C.; Li, L.; Yu, L.; El~Kholy, A.; Ahmed, F.; Gan, Z.; Cheng, Y.; and Liu, J. 2020{\natexlab{b}}.
\newblock Uniter: Universal image-text representation learning.
\newblock In \emph{European conference on computer vision}, 104--120. Springer.

\bibitem[{Chua et~al.(2009)Chua, Tang, Hong, Li, Luo, and Zheng}]{chua2009nus}
Chua, T.-S.; Tang, J.; Hong, R.; Li, H.; Luo, Z.; and Zheng, Y. 2009.
\newblock Nus-wide: a real-world web image database from national university of singapore.
\newblock In \emph{Proceedings of the ACM international conference on image and video retrieval}, 1--9.

\bibitem[{Guo et~al.(2022)Guo, Mao, Wei, and Huang}]{guo2022intra}
Guo, J.-N.; Mao, X.-L.; Wei, W.; and Huang, H. 2022.
\newblock Intra-category aware hierarchical supervised document hashing.
\newblock \emph{IEEE Transactions on Knowledge and Data Engineering}, 35(6): 6003--6013.

\bibitem[{Huiskes, Thomee, and Lew(2010)}]{huiskes2010new}
Huiskes, M.~J.; Thomee, B.; and Lew, M.~S. 2010.
\newblock New trends and ideas in visual concept detection: The mir flickr retrieval evaluation initiative.
\newblock In \emph{Proceedings of the international conference on Multimedia information retrieval}, 527--536.

\bibitem[{Iscen et~al.(2022)Iscen, Valmadre, Arnab, and Schmid}]{iscen2022learning}
Iscen, A.; Valmadre, J.; Arnab, A.; and Schmid, C. 2022.
\newblock Learning with neighbor consistency for noisy labels.
\newblock In \emph{Proceedings of the IEEE/CVF Conference on Computer Vision and Pattern Recognition}, 4672--4681.

\bibitem[{Jiang et~al.(2024)Jiang, Zhang, Bai, Wang, and Meng}]{jiang2024more}
Jiang, G.; Zhang, J.; Bai, X.; Wang, W.; and Meng, D. 2024.
\newblock Which Is More Effective in Label Noise Cleaning, Correction or Filtering?
\newblock In \emph{Proceedings of the AAAI Conference on Artificial Intelligence}, volume~38, 12866--12873.

\bibitem[{Jiang and Li(2017)}]{jiang2017deep}
Jiang, Q.-Y.; and Li, W.-J. 2017.
\newblock Deep cross-modal hashing.
\newblock In \emph{Proceedings of the IEEE conference on computer vision and pattern recognition}, 3232--3240.

\bibitem[{Kim et~al.(2021)Kim, Ko, Choi, Yun et~al.}]{kim2021fine}
Kim, T.; Ko, J.; Choi, J.; Yun, S.-Y.; et~al. 2021.
\newblock Fine samples for learning with noisy labels.
\newblock \emph{Advances in Neural Information Processing Systems}, 34: 24137--24149.

\bibitem[{Li et~al.(2024)Li, Shu, Yu, and Wu}]{li2024robust}
Li, L.; Shu, Z.; Yu, Z.; and Wu, X.-J. 2024.
\newblock Robust online hashing with label semantic enhancement for cross-modal retrieval.
\newblock \emph{Pattern Recognition}, 145: 109972.

\bibitem[{Lin et~al.(2014)Lin, Maire, Belongie, Hays, Perona, Ramanan, Doll{\'a}r, and Zitnick}]{lin2014microsoft}
Lin, T.-Y.; Maire, M.; Belongie, S.; Hays, J.; Perona, P.; Ramanan, D.; Doll{\'a}r, P.; and Zitnick, C.~L. 2014.
\newblock Microsoft coco: Common objects in context.
\newblock In \emph{Computer Vision--ECCV 2014: 13th European Conference, Zurich, Switzerland, September 6-12, 2014, Proceedings, Part V 13}, 740--755. Springer.

\bibitem[{Liu, He, and Lang(2014)}]{liu2014multiple}
Liu, X.; He, J.; and Lang, B. 2014.
\newblock Multiple feature kernel hashing for large-scale visual search.
\newblock \emph{Pattern Recognition}, 47(2): 748--757.

\bibitem[{Lu et~al.(2020)Lu, Liu, Nie, Chang, and Zhang}]{lu2020semantic}
Lu, X.; Liu, L.; Nie, L.; Chang, X.; and Zhang, H. 2020.
\newblock Semantic-driven interpretable deep multi-modal hashing for large-scale multimedia retrieval.
\newblock \emph{IEEE Transactions on Multimedia}, 23: 4541--4554.

\bibitem[{Lu et~al.(2019{\natexlab{a}})Lu, Zhu, Cheng, Li, Nie, and Zhang}]{lu2019flexible}
Lu, X.; Zhu, L.; Cheng, Z.; Li, J.; Nie, X.; and Zhang, H. 2019{\natexlab{a}}.
\newblock Flexible online multi-modal hashing for large-scale multimedia retrieval.
\newblock In \emph{Proceedings of the 27th ACM international conference on multimedia}, 1129--1137.

\bibitem[{Lu et~al.(2019{\natexlab{b}})Lu, Zhu, Cheng, Nie, and Zhang}]{lu2019online}
Lu, X.; Zhu, L.; Cheng, Z.; Nie, L.; and Zhang, H. 2019{\natexlab{b}}.
\newblock Online multi-modal hashing with dynamic query-adaption.
\newblock In \emph{Proceedings of the 42nd international ACM SIGIR conference on research and development in information retrieval}, 715--724.

\bibitem[{Lu et~al.(2019{\natexlab{c}})Lu, Zhu, Li, Zhang, and Shen}]{lu2019efficient}
Lu, X.; Zhu, L.; Li, J.; Zhang, H.; and Shen, H.~T. 2019{\natexlab{c}}.
\newblock Efficient supervised discrete multi-view hashing for large-scale multimedia search.
\newblock \emph{IEEE Transactions on Multimedia}, 22(8): 2048--2060.

\bibitem[{Lu et~al.(2021)Lu, Zhu, Liu, Nie, and Zhang}]{lu2021graph}
Lu, X.; Zhu, L.; Liu, L.; Nie, L.; and Zhang, H. 2021.
\newblock Graph convolutional multi-modal hashing for flexible multimedia retrieval.
\newblock In \emph{Proceedings of the 29th ACM international conference on multimedia}, 1414--1422.

\bibitem[{Shen et~al.(2023)Shen, Chen, Pan, Liu, and Zheng}]{shen2023graph}
Shen, X.; Chen, Y.; Pan, S.; Liu, W.; and Zheng, Y. 2023.
\newblock Graph convolutional incomplete multi-modal hashing.
\newblock In \emph{Proceedings of the 31st ACM international conference on multimedia}, 7029--7037.

\bibitem[{Shen et~al.(2018)Shen, Shen, Liu, Yuan, Liu, and Sun}]{shen2018multiview}
Shen, X.; Shen, F.; Liu, L.; Yuan, Y.-H.; Liu, W.; and Sun, Q.-S. 2018.
\newblock Multiview discrete hashing for scalable multimedia search.
\newblock \emph{ACM Transactions on Intelligent Systems and Technology (TIST)}, 9(5): 1--21.

\bibitem[{Shen et~al.(2015)Shen, Shen, Sun, and Yuan}]{shen2015multi}
Shen, X.; Shen, F.; Sun, Q.-S.; and Yuan, Y.-H. 2015.
\newblock Multi-view latent hashing for efficient multimedia search.
\newblock In \emph{Proceedings of the 23rd ACM international conference on Multimedia}, 831--834.

\bibitem[{Shu et~al.(2019)Shu, Xie, Yi, Zhao, Zhou, Xu, and Meng}]{shu2019meta}
Shu, J.; Xie, Q.; Yi, L.; Zhao, Q.; Zhou, S.; Xu, Z.; and Meng, D. 2019.
\newblock Meta-weight-net: Learning an explicit mapping for sample weighting.
\newblock \emph{Advances in neural information processing systems}, 32.

\bibitem[{Simonyan and Zisserman(2014)}]{simonyan2014very}
Simonyan, K.; and Zisserman, A. 2014.
\newblock Very deep convolutional networks for large-scale image recognition.
\newblock \emph{arXiv preprint arXiv:1409.1556}.

\bibitem[{Song et~al.(2020)Song, Dai, Raskutti, and Barber}]{song2020convex}
Song, H.; Dai, R.; Raskutti, G.; and Barber, R.~F. 2020.
\newblock Convex and non-Convex approaches for statistical inference with class-conditional noisy labels.
\newblock \emph{Journal of Machine Learning Research}, 21(168): 1--58.

\bibitem[{Song et~al.(2022)Song, Kim, Park, Shin, and Lee}]{song2022learning}
Song, H.; Kim, M.; Park, D.; Shin, Y.; and Lee, J.-G. 2022.
\newblock Learning from noisy labels with deep neural networks: A survey.
\newblock \emph{IEEE transactions on neural networks and learning systems}, 34(11): 8135--8153.

\bibitem[{Song et~al.(2013)Song, Yang, Huang, Shen, and Luo}]{song2013effective}
Song, J.; Yang, Y.; Huang, Z.; Shen, H.~T.; and Luo, J. 2013.
\newblock Effective multiple feature hashing for large-scale near-duplicate video retrieval.
\newblock \emph{IEEE Transactions on Multimedia}, 15(8): 1997--2008.

\bibitem[{Sun et~al.(2022)Sun, Wang, Luo, Zhang, Xiang, Chen, and Hua}]{sun2022heart}
Sun, J.; Wang, H.; Luo, X.; Zhang, S.; Xiang, W.; Chen, C.; and Hua, X.-S. 2022.
\newblock Heart: Towards effective hash codes under label noise.
\newblock In \emph{Proceedings of the 30th ACM International Conference on Multimedia}, 366--375.

\bibitem[{Sung et~al.(2018)Sung, Yang, Zhang, Xiang, Torr, and Hospedales}]{sung2018learning}
Sung, F.; Yang, Y.; Zhang, L.; Xiang, T.; Torr, P.~H.; and Hospedales, T.~M. 2018.
\newblock Learning to compare: Relation network for few-shot learning.
\newblock In \emph{Proceedings of the IEEE conference on computer vision and pattern recognition}, 1199--1208.

\bibitem[{Tan et~al.(2022)Tan, Zhu, Guan, Li, and Cheng}]{tan2022bit}
Tan, W.; Zhu, L.; Guan, W.; Li, J.; and Cheng, Z. 2022.
\newblock Bit-aware semantic transformer hashing for multi-modal retrieval.
\newblock In \emph{Proceedings of the 45th international ACM SIGIR conference on research and development in information retrieval}, 982--991.

\bibitem[{Tan et~al.(2023)Tan, Zhu, Li, Zhang, and Zhang}]{tan2023partial}
Tan, W.; Zhu, L.; Li, J.; Zhang, Z.; and Zhang, H. 2023.
\newblock Partial multi-modal hashing via neighbor-aware completion learning.
\newblock \emph{IEEE Transactions on Multimedia}, 25: 8499--8510.

\bibitem[{Tu et~al.(2021{\natexlab{a}})Tu, Ji, Luo, Shi, Huang, Duan, and Mao}]{tu2021hashing}
Tu, R.-C.; Ji, L.; Luo, H.; Shi, B.; Huang, H.-Y.; Duan, N.; and Mao, X.-L. 2021{\natexlab{a}}.
\newblock Hashing based efficient inference for image-text matching.
\newblock In \emph{Findings of the Association for Computational Linguistics: ACL-IJCNLP 2021}, 743--752.

\bibitem[{Tu et~al.(2023{\natexlab{a}})Tu, Jiang, Lin, Cai, Tian, Wang, and Liu}]{tu2023unsupervised}
Tu, R.-C.; Jiang, J.; Lin, Q.; Cai, C.; Tian, S.; Wang, H.; and Liu, W. 2023{\natexlab{a}}.
\newblock Unsupervised cross-modal hashing with modality-interaction.
\newblock \emph{IEEE Transactions on Circuits and Systems for Video Technology}, 33(9): 5296--5308.

\bibitem[{Tu, Mao, and Wei(2020)}]{tu2020mls3rduh}
Tu, R.-C.; Mao, X.; and Wei, W. 2020.
\newblock MLS3RDUH: Deep Unsupervised Hashing via Manifold based Local Semantic Similarity Structure Reconstructing.
\newblock In \emph{IJCAI}, 3466--3472.

\bibitem[{Tu et~al.(2018)Tu, Mao, Feng, Bian, and Ying}]{tu2018object}
Tu, R.-C.; Mao, X.-L.; Feng, B.-S.; Bian, B.-B.; and Ying, Y.-s. 2018.
\newblock Object detection based deep unsupervised hashing.
\newblock \emph{arXiv preprint arXiv:1811.09822}.

\bibitem[{Tu et~al.(2021{\natexlab{b}})Tu, Mao, Guo, Wei, and Huang}]{tu2021partial}
Tu, R.-C.; Mao, X.-L.; Guo, J.-N.; Wei, W.; and Huang, H. 2021{\natexlab{b}}.
\newblock Partial-softmax loss based deep hashing.
\newblock In \emph{Proceedings of the Web Conference 2021}, 2869--2878.

\bibitem[{Tu et~al.(2023{\natexlab{b}})Tu, Mao, Ji, Wei, and Huang}]{tu2023data}
Tu, R.-C.; Mao, X.-L.; Ji, W.; Wei, W.; and Huang, H. 2023{\natexlab{b}}.
\newblock Data-aware proxy hashing for cross-modal retrieval.
\newblock In \emph{Proceedings of the 46th International ACM SIGIR Conference on Research and Development in Information Retrieval}, 686--696.

\bibitem[{Tu et~al.(2021{\natexlab{c}})Tu, Mao, Kong, Shao, Li, Wei, and Huang}]{tu2021weighted}
Tu, R.-C.; Mao, X.-L.; Kong, C.; Shao, Z.; Li, Z.-L.; Wei, W.; and Huang, H. 2021{\natexlab{c}}.
\newblock Weighted gaussian loss based hamming hashing.
\newblock In \emph{Proceedings of the 29th ACM International Conference on Multimedia}, 3409--3417.

\bibitem[{Tu et~al.(2024)Tu, Mao, Liu, Wei, Huang et~al.}]{tu2024similarity}
Tu, R.-C.; Mao, X.-L.; Liu, J.; Wei, W.; Huang, H.; et~al. 2024.
\newblock Similarity Transitivity Broken-Aware Multi-Modal Hashing.
\newblock \emph{IEEE Transactions on Knowledge and Data Engineering}.

\bibitem[{Tu et~al.(2022)Tu, Mao, Tu, Bian, Cai, Wang, Wei, and Huang}]{tu2022deep}
Tu, R.-C.; Mao, X.-L.; Tu, R.-X.; Bian, B.; Cai, C.; Wang, H.; Wei, W.; and Huang, H. 2022.
\newblock Deep cross-modal proxy hashing.
\newblock \emph{IEEE Transactions on Knowledge and Data Engineering}, 35(7): 6798--6810.

\bibitem[{Wang et~al.(2023)Wang, Jiang, Sun, Zhang, Chen, Hua, and Luo}]{wang2023dior}
Wang, H.; Jiang, H.; Sun, J.; Zhang, S.; Chen, C.; Hua, X.-S.; and Luo, X. 2023.
\newblock Dior: Learning to hash with label noise via dual partition and contrastive learning.
\newblock \emph{IEEE Transactions on Knowledge and Data Engineering}.

\bibitem[{Wei et~al.(2022)Wei, Sun, Lu, and Yin}]{wei2022self}
Wei, Q.; Sun, H.; Lu, X.; and Yin, Y. 2022.
\newblock Self-filtering: A noise-aware sample selection for label noise with confidence penalization.
\newblock In \emph{European Conference on Computer Vision}, 516--532. Springer.

\bibitem[{Wu et~al.(2021{\natexlab{a}})Wu, Zhu, Xie, Zhang, and Zhang}]{wu2021multi}
Wu, X.; Zhu, L.; Xie, L.; Zhang, Z.; and Zhang, H. 2021{\natexlab{a}}.
\newblock Multi-modal discrete tensor decomposition hashing for efficient multimedia retrieval.
\newblock \emph{Neurocomputing}, 465: 1--14.

\bibitem[{Wu et~al.(2022)Wu, Luo, Zhan, Ding, Chen, and Xu}]{wu2022online}
Wu, X.-M.; Luo, X.; Zhan, Y.-W.; Ding, C.-L.; Chen, Z.-D.; and Xu, X.-S. 2022.
\newblock Online enhanced semantic hashing: Towards effective and efficient retrieval for streaming multi-modal data.
\newblock In \emph{Proceedings of the AAAI conference on artificial intelligence}, volume~36, 4263--4271.

\bibitem[{Wu et~al.(2021{\natexlab{b}})Wu, Shu, Xie, Zhao, and Meng}]{wu2021learning}
Wu, Y.; Shu, J.; Xie, Q.; Zhao, Q.; and Meng, D. 2021{\natexlab{b}}.
\newblock Learning to purify noisy labels via meta soft label corrector.
\newblock In \emph{Proceedings of the AAAI Conference on Artificial Intelligence}, volume~35, 10388--10396.

\bibitem[{Xie et~al.(2017)Xie, Shen, Han, Zhu, and Shao}]{xie2017dynamic}
Xie, L.; Shen, J.; Han, J.; Zhu, L.; and Shao, L. 2017.
\newblock Dynamic Multi-View Hashing for Online Image Retrieval.
\newblock In \emph{IJCAI}, volume~78, 122.

\bibitem[{Yan et~al.(2020)Yan, Gong, Wei, and Gao}]{yan2020deep}
Yan, C.; Gong, B.; Wei, Y.; and Gao, Y. 2020.
\newblock Deep multi-view enhancement hashing for image retrieval.
\newblock \emph{IEEE Transactions on Pattern Analysis and Machine Intelligence}, 43(4): 1445--1451.

\bibitem[{Yang et~al.(2022)Yang, Yao, Liu, and Deng}]{yang2022mutual}
Yang, E.; Yao, D.; Liu, T.; and Deng, C. 2022.
\newblock Mutual quantization for cross-modal search with noisy labels.
\newblock In \emph{Proceedings of the IEEE/CVF Conference on Computer Vision and Pattern Recognition}, 7551--7560.

\bibitem[{Yang, Shi, and Xu(2017)}]{yang2017discrete}
Yang, R.; Shi, Y.; and Xu, X.-S. 2017.
\newblock Discrete multi-view hashing for effective image retrieval.
\newblock In \emph{Proceedings of the 2017 ACM on international conference on multimedia retrieval}, 175--183.

\bibitem[{Yu et~al.(2022)Yu, Huang, Li, Shu, and Zhu}]{yu2022hadamard}
Yu, J.; Huang, W.; Li, Z.; Shu, Z.; and Zhu, L. 2022.
\newblock Hadamard matrix-guided multi-modal hashing for multi-modal retrieval.
\newblock \emph{Digital Signal Processing}, 130: 103743.

\bibitem[{Yuan et~al.(2020)Yuan, Wang, Zhang, Tay, Jie, Liu, and Feng}]{yuan2020central}
Yuan, L.; Wang, T.; Zhang, X.; Tay, F.~E.; Jie, Z.; Liu, W.; and Feng, J. 2020.
\newblock Central similarity quantization for efficient image and video retrieval.
\newblock In \emph{Proceedings of the IEEE/CVF conference on computer vision and pattern recognition}, 3083--3092.

\bibitem[{Zhang, Peng, and Yuan(2018)}]{zhang2018unsupervised}
Zhang, J.; Peng, Y.; and Yuan, M. 2018.
\newblock Unsupervised generative adversarial cross-modal hashing.
\newblock In \emph{Proceedings of the AAAI conference on artificial intelligence}, volume~32.

\bibitem[{Zhang, Lai, and Feng(2018)}]{zhang2018attention}
Zhang, X.; Lai, H.; and Feng, J. 2018.
\newblock Attention-aware deep adversarial hashing for cross-modal retrieval.
\newblock In \emph{Proceedings of the European conference on computer vision (ECCV)}, 591--606.

\bibitem[{Zheng et~al.(2020{\natexlab{a}})Zheng, Zhu, Cheng, Li, and Liu}]{zheng2020adaptive}
Zheng, C.; Zhu, L.; Cheng, Z.; Li, J.; and Liu, A.-A. 2020{\natexlab{a}}.
\newblock Adaptive partial multi-view hashing for efficient social image retrieval.
\newblock \emph{IEEE Transactions on Multimedia}, 23: 4079--4092.

\bibitem[{Zheng et~al.(2019)Zheng, Zhu, Lu, Li, Cheng, and Zhang}]{zheng2019fast}
Zheng, C.; Zhu, L.; Lu, X.; Li, J.; Cheng, Z.; and Zhang, H. 2019.
\newblock Fast discrete collaborative multi-modal hashing for large-scale multimedia retrieval.
\newblock \emph{IEEE Transactions on Knowledge and Data Engineering}, 32(11): 2171--2184.

\bibitem[{Zheng et~al.(2020{\natexlab{b}})Zheng, Zhu, Zhang, and Zhang}]{zheng2020efficient}
Zheng, C.; Zhu, L.; Zhang, S.; and Zhang, H. 2020{\natexlab{b}}.
\newblock Efficient parameter-free adaptive multi-modal hashing.
\newblock \emph{IEEE Signal Processing Letters}, 27: 1270--1274.

\bibitem[{Zheng et~al.(2024)Zheng, Zhu, Zhang, Duan, and Lu}]{zheng2024lcemh}
Zheng, C.; Zhu, L.; Zhang, Z.; Duan, W.; and Lu, W. 2024.
\newblock LCEMH: Label Correlation Enhanced Multi-modal Hashing for efficient multi-modal retrieval.
\newblock \emph{Information Sciences}, 659: 120064.

\bibitem[{Zheng et~al.(2022)Zheng, Zhu, Zhang, Li, and Yu}]{zheng2022efficient}
Zheng, C.; Zhu, L.; Zhang, Z.; Li, J.; and Yu, X. 2022.
\newblock Efficient semi-supervised multimodal hashing with importance differentiation regression.
\newblock \emph{IEEE Transactions on Image Processing}, 31: 5881--5892.

\bibitem[{Zheng et~al.(2020{\natexlab{c}})Zheng, Wu, Goswami, Goswami, Metaxas, and Chen}]{zheng2020error}
Zheng, S.; Wu, P.; Goswami, A.; Goswami, M.; Metaxas, D.; and Chen, C. 2020{\natexlab{c}}.
\newblock Error-bounded correction of noisy labels.
\newblock In \emph{International Conference on Machine Learning}, 11447--11457. PMLR.

\bibitem[{Zhu et~al.(2020)Zhu, Lu, Cheng, Li, and Zhang}]{zhu2020deep}
Zhu, L.; Lu, X.; Cheng, Z.; Li, J.; and Zhang, H. 2020.
\newblock Deep collaborative multi-view hashing for large-scale image search.
\newblock \emph{IEEE Transactions on Image Processing}, 29: 4643--4655.

\bibitem[{Zhu et~al.(2021)Zhu, Zheng, Lu, Cheng, Nie, and Zhang}]{zhu2021efficient}
Zhu, L.; Zheng, C.; Lu, X.; Cheng, Z.; Nie, L.; and Zhang, H. 2021.
\newblock Efficient multi-modal hashing with online query adaption for multimedia retrieval.
\newblock \emph{ACM Transactions on Information Systems (TOIS)}, 40(2): 1--36.

\end{thebibliography}

\end{document}